\documentclass[letterpaper]{article} % DO NOT CHANGE THIS
\usepackage{aaai2026}  % DO NOT CHANGE THIS
\usepackage{times}  % DO NOT CHANGE THIS
\usepackage{helvet}  % DO NOT CHANGE THIS
\usepackage{courier}  % DO NOT CHANGE THIS
\usepackage[hyphens]{url}  % DO NOT CHANGE THIS
\usepackage{graphicx} % DO NOT CHANGE THIS
\usepackage{natbib}  % DO NOT CHANGE THIS AND DO NOT ADD ANY OPTIONS TO IT
\usepackage{caption} % DO NOT CHANGE THIS AND DO NOT ADD ANY OPTIONS TO IT
\usepackage{algorithm}
\usepackage{algorithmic}
\usepackage{url}            % simple URL typesetting
\usepackage{booktabs}       % professional-quality tables
\usepackage{amsfonts}       % blackboard math symbols
\usepackage{nicefrac}       % compact symbols for 1/2, etc.
\usepackage{microtype}      % microtypography
\usepackage{amssymb}
\usepackage{tabularx}
\usepackage{pifont}
\usepackage{subcaption}
\usepackage{newfloat}
\usepackage{listings}
\DeclareCaptionStyle{ruled}{labelfont=normalfont,labelsep=colon,strut=off} % DO NOT CHANGE THIS
\floatstyle{ruled}
\newfloat{listing}{tb}{lst}{}
\floatname{listing}{Listing}
\usepackage{enumitem}
\usepackage[skins]{tcolorbox}
\usepackage{pifont}
\definecolor{myblue}{RGB}{44, 175,250}
\definecolor{mybluex}{RGB}{162, 220,250}
\definecolor{myred}{RGB}{238, 96,137}
\usepackage{amsmath}
\tcbuselibrary{breakable}

\usepackage{newfloat}
\usepackage{listings}
\DeclareCaptionStyle{ruled}{labelfont=normalfont,labelsep=colon,strut=off} % DO NOT CHANGE THIS
\floatstyle{ruled}
\newfloat{listing}{tb}{lst}{}
\floatname{listing}{Listing}

\title{Mediator-Guided Multi-Agent Collaboration among Open-Source Models for Medical Decision-Making}
\author {
    Kaitao Chen\textsuperscript{\rm 1,2}, 
    Mianxin Liu\textsuperscript{\rm 2}\Corresponding, 
    Daoming Zong\textsuperscript{\rm 3},
    Chaoyue Ding\textsuperscript{\rm 1}, 
    Shaohao Rui\textsuperscript{\rm 2}, 
    Yankai Jiang\textsuperscript{\rm 2}, \\
    Mu Zhou\textsuperscript{\rm 4}\Corresponding, 
    Xiaosong Wang\textsuperscript{\rm 2}
}
\affiliations {
    \textsuperscript{\rm 1}Fudan University\\
    \textsuperscript{\rm 2}Shanghai AI Laboratory\\
    \textsuperscript{\rm 3}East China Normal University\\
    \textsuperscript{\rm 4}Rutgers University\\
    chenkaitao@pjlab.org.cn, liumianxin@pjlab.org.cn, mu.zhou2010@gmail.com
}

\usepackage{bibentry}
\begin{document}

\maketitle

\begin{abstract}
Complex medical decision-making involves cooperative workflows operated by different clinicians. Designing AI multi-agent systems can expedite and augment human-level clinical decision-making. Existing multi-agent researches primarily focus on language-only tasks, yet their extension to multimodal scenarios remains challenging. A blind combination of diverse vision-language models (VLMs) can amplify an erroneous outcome interpretation. VLMs in general are less capable in instruction following and importantly self-reflection, compared to large language models (LLMs) of comparable sizes. This disparity largely constrains VLMs' ability in cooperative workflows. In this study, we propose MedOrch, a mediator-guided multi-agent collaboration framework for medical multimodal decision-making. MedOrch employs an LLM-based mediator agent that enables multiple VLM-based expert agents to exchange and reflect on their outputs towards collaboration. We utilize multiple open-source general-purpose and domain-specific VLMs instead of costly GPT-series models, revealing the strength of heterogeneous models. We show that the collaboration within distinct VLM-based agents can surpass the capabilities of any individual agent. We validate our approach on five medical vision question answering benchmarks, demonstrating superior collaboration performance without model training. Our findings underscore the value of mediator-guided multi-agent collaboration in advancing medical multimodal intelligence. 
% Our code will be made publicly available.

\end{abstract}

\section{Introduction}
Complex clinical decision-making hinges on a comprehensive cognitive ability from medical vision, language, and domain knowledge derived from human experts. Using large language models (LLMs) alone can only support language-level assessment \cite{llmmedicine1, llmmedicine2}. The exploration of vision-language models (VLMs) opens the possibility to clinical multimodal decision-making driven by model advances \cite{gpt-4vmedical, llavamed, huatuogptvision}. Incorporating a joint strength of these models can enable a collaborative framework to understand and inform high-quality decision-making.

Emerging multi-agent systems have garnered increasing attention for enhancing clinical decision-making \cite{zhang2024data,proagent}. In contrast to a single agent, a multi-agent system supports collaboration among different agents. Akin to the real-world clinical workflow to glean various diagnostic opinions, multi-agent systems can mirror the patient-care coordination from diverse perspectives \cite{medaide}.

% However, current multi-agent systems are often restricted in handling language-only tasks. 

A multi-agent system often follows a voting or judging-based strategy \cite{voting}, allowing multiple agents to contribute to the decision-making process. To promote a deep interaction, it is intuitive to establish competitive or collaborative agent roles \cite{moa,camel}. Collaborative frameworks from Mixture of Agents (MoA) \cite{moa}, CAMEL \cite{camel}, and ProAgent~\cite{proagent} highlight the model's cognitive preference to reveal complex cooperative reasoning. In medicine, Multi-expert Prompting \cite{long} and MedAgents \cite{medagents} are examples for building a role-playing multi-agent framework. However, these multi-agent systems are largely restricted in handling language-only tasks. Although MDagents \cite{mdagents} seeks a multimodal question answering, it uses the GPT-4V for the small-scale experiment on vision question answering (VQA) datasets (e.g., 50 samples per dataset) due to the prohibitive cost. Meanwhile, several open-source medical VLMs have begun outperforming general-purpose counterparts like GPT-4o in medical VQA \cite{MedMax}, motivating us to design a multi-VLM-agent system based on open-source models without API cost. This raises an intriguing question: \textbf{Can open-source multimodal agents with heterogeneous abilities develop effective collaboration strategies for clinical decision-making}?

Realistic barriers emerge in constructing diverse agents and their collaboration when building a multimodal decision-making system. General-purpose models provide broad knowledge coverage, while domain-specific models excel in capturing nuances in specialized scenarios. A blind combination of these models can generate erroneous interpretations and propagate via agent interactions. In addition, VLM's instruction following and importantly self-reflection capabilities remain inferior to those of LLMs in similar sizes~\cite{medxpertqa}. This remarkable gap limits the ability of VLMs to engage in meaningful reflection, an essential capability in the progress of multi-agent collaboration. Given the heterogeneous nature and remarkable difference in capability of VLM agents, it becomes further challenging to coordinate their interactions effectively. 

\begin{figure*}
    \centering
    \includegraphics[width=\linewidth]{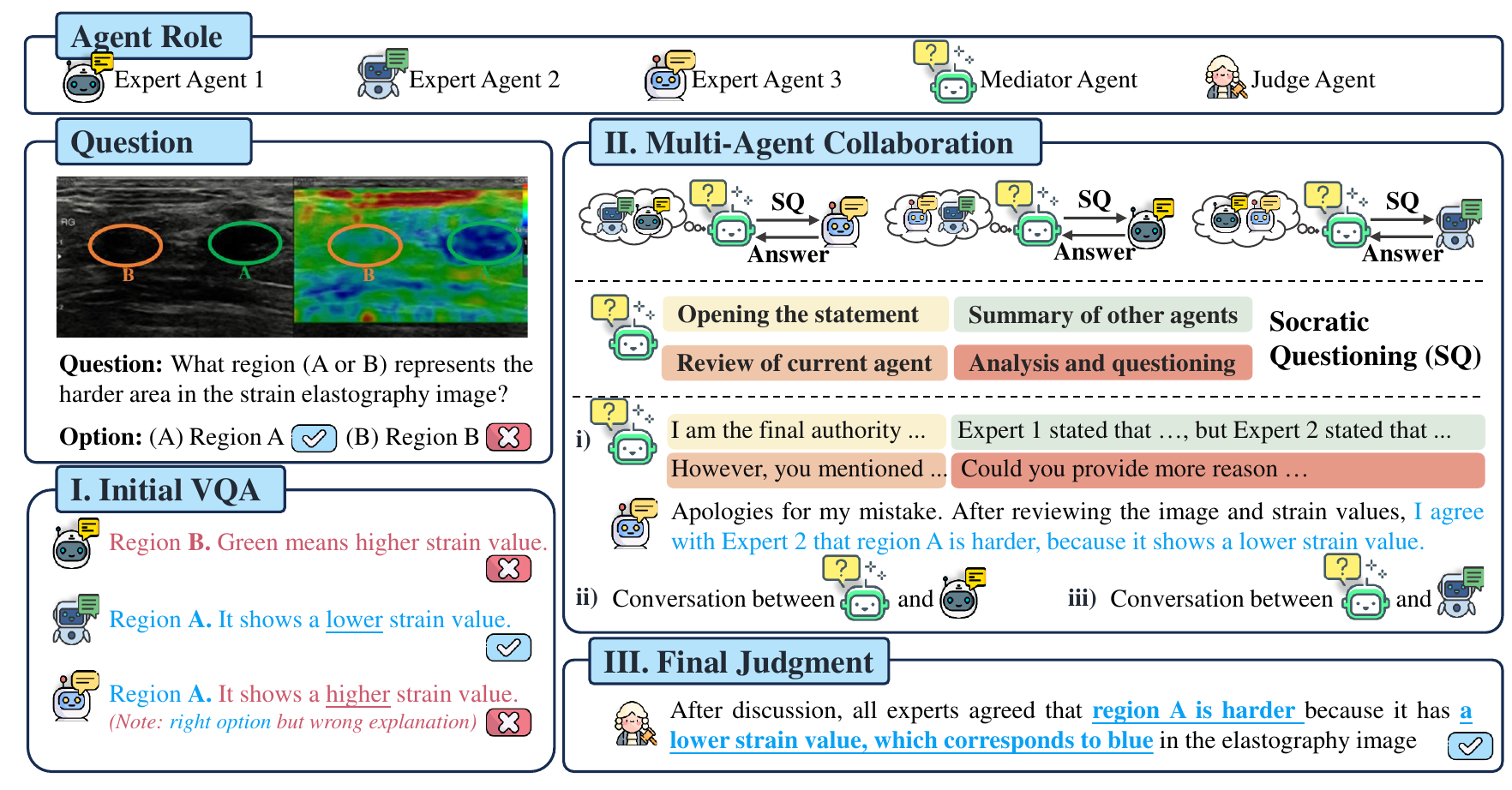}
    \caption{The mediator-guided multi-agent collaboration framework for medical VQA. In the initial stage, multiple VLM expert agents independently generate preliminary answers to a given question. To promote deeper interaction, the mediator agent synthesizes the information and formulates Socratic questions for expert agents. Subsequently, the relevant expert agents reflect on the question and generate refined responses accordingly. Finally, a judge agent analyzes the dialogues between the mediator and expert agents and achieves a systematic output.}
    \label{fig:arch}
\vspace{-0.3cm}
\end{figure*}

We draw inspiration from social theory, where Socratic questioning \cite{Socratic} fosters disciplined and reflective dialogue through a sequence of structured questions. Such an approach encourages critical reflection, evidence scrutiny, and clarification of conceptual ambiguities \cite{socratic2}. Translating this insight to AI collaboration, we propose \textbf{MedOrch}, a mediator-guided multi-agent collaboration framework. MedOrch incorporates a dedicated LLM-based mediator agent that orchestrates the interaction and reflection among multiple VLM-based expert agents to support collaborative medical decision-making. Specifically, we construct the VLM-based expert agents using open-source general-purpose and domain-specific VLMs, leveraging their complementary strengths, rather than using cost-prohibitive GPT-series models. To start, in Figure~\ref{fig:arch}, multiple VLM-based expert agents first generate responses to a given question. The mediator agent then synthesizes the information and formulates Socratic questions to prompt deeper reasoning from the VLM-based expert agent. Next, relevant VLM-based expert agents rethink and generate their responses based on the question of the mediator agent. Finally, a judge agent integrates dialogues and provides a final judgment to support medical decision-making.

The main contributions of this study are: \textbf{(i)} We propose MedOrch as a novel multi-agent collaboration for improved medical decision-making. We design the mediator agent to guide the interaction of open-source VLM-based expert agents. \textbf{(ii)} Our design emphasizes joint abilities gained from distinct open-source general-purpose and domain-specific VLMs to enable multimodal decision-making without prohibitive API cost. It is noteworthy that our multi-agent system outperforms the best of multiple single-agent models, even in the presence of misleading inputs from certain agents. \textbf{(iii)} We conduct comprehensive evaluations on five benchmarks. Our approach achieves strong performance, demonstrating its potential efficiency in medical multimodal scenarios.

% MDAgents \cite{mdagents} simulates real-world medical decision models through dynamic collaboration of AI agents, tailored to task complexity, and offering practical solutions for medical decision-making processes.

\section{Related Work}
\subsection{Multi-Agent AI System} 
While LLMs as individual agents \cite{qwen2.5,llama3} reveal language understanding, evident challenges persist in complex reasoning, interpretation, and decision-making \cite{AgentSurvey1, AgentSurvey2}. As a natural progression, building a multi-agent system requires us to cultivate collective intelligence through mechanisms of cooperation or competition \cite{mdagents, moa, debate, proagent}. 

Cooperative learning fosters the collective knowledge sharing and joint decision-making via agent voting \cite{voting}, interdisciplinary collaboration \cite{mdagents}, and group discussions \cite{reconcile}. For instance, ProAgent~\cite{proagent} is a multi-agent reinforcement learning framework leveraging policy sharing and inter-agent communication. Similarly, CORY \cite{cory} emulated human reflective group learning by implementing cycles of proposal, negotiation, reflection, and revision in programming and mathematical tasks. CAMEL \cite{camel} adopted a role-playing collaborative framework to allow autonomous dialogue and multi-step coordination among agents, yet it falls into role confusion and instruction redundancy. In the meantime, competitive approach stimulates reasoning and creativity through conflict, adversarial interactions, and debate \cite{debate, MAD}. The Multi-Agent Debate framework \cite{debate} mitigates the overconfidence in LLMs by introducing viewpoint conflict to promote deeper reasoning. Similarly, MAD~\cite{MAD} integrates diverse perspectives through iterative critique and revision to enhance mathematical reasoning and factual verification. MoA \cite{moa} introduces a multi-agent collaboration by organizing LLMs into a hierarchical architecture in language-only tasks. However, extending MoA into multimodal scenarios is problematic, since VLMs may amplify misinterpretations from other agents, ultimately hindering the effectiveness of the collaborative workflow~\cite{mdagents}.

% Beyond mechanisms of cooperation or competition, multi-agent systems increasingly emphasize the integration of external resources to further strengthen reasoning and execution capabilities \cite{knowagent, wkm,eto, self-reflection, critic}. Incorporating external structured knowledge bases and expert experience could help agent planning. For example, KnowAgent \cite{knowagent} combined action knowledge and decision constraints to mitigate hallucinations during LLM-based planning, while WKM \cite{wkm} fused global priors with dynamic local information to guide efficient decision-making in complex scenarios. On the other hand, external tools, evaluator feedback, and human evaluations play crucial roles in iterative agent improvement to dynamically evolve and refine agent behavior \cite{eto, self-reflection, critic}.

% The systematic integration of external resources not only enhances decision accuracy in individual interactions but also significantly improves adaptability and generalization over the course of long-term evolution, laying a robust foundation for the development of more resilient and intelligent multi-agent systems.

\subsection{Multi-Agent in Healthcare}
Multi-agent systems in healthcare can be used to advance clinical knowledge sharing and workflow automation \cite{zhang2024data}. Emerging efforts have focused on (a) role configuration \cite{long,MDTeamGPT,pandey2024advancing}, (b) task decomposition \cite{pathgen,medaide}, and (c) workflow simulation \cite{medco,li}. First, role configuration starts with multi-expert prompting \cite{long} to generate expert roles in a zero-shot manner. MDTeamGPT~\cite{MDTeamGPT} develops a multi-agent system for multidisciplinary medical consultations, incorporating a primary physician mechanism and a residual discussion structure. Second, task decomposition utilizes multiple role-playing agents to improve the task performance. For instance, PathGen \cite{pathgen} enhances the quality of pathological captions through role-based agents. MEDAID \cite{medaide} decomposes complex medical queries through query rewriting and context encoders. Third, workflow simulation frameworks are designed to mimic insights in clinical environments. MEDCO \cite{medco} simulates the collaboration of patients, medical experts, and doctors, providing personalized and precise clinical case training. In addition, agent hospital \cite{li} simulates a hospital system that integrates domain-specific knowledge bases to automate data generation and address specific tasks. 

%Pandey et al. \cite{pandey2024advancing} explored the application of multi-agent systems in medical procedure necessity determination, proposing a multi-agent system that reasons by comparing clinical guidelines with patient records to complete pre-authorization tasks. 

The above multi-agent studies are primarily driven by single-modal and close-source LLMs. Multi-agent systems remain largely underexplored in multimodal decision-making scenarios, especially under the capacity of open-source models. Although MDagents \cite{mdagents} leverages the multi-agent to address medical VQA tasks, its reliance on a limited-scale experiment (only 50 samples per dataset) and the high cost of GPT-4V significantly restricts its clinical applicability. In contrast, our study focuses on complex medical multimodal scenarios by integrating different open-source vision-language and language-only models, as well as by combining general-purpose and domain-specific models, to enable practical clinical multimodal decision-making. 

% Table \ref{difference} compares existing multi-agent approaches from multiple dimensions.

\section{Method} 
\subsection{Overview}
MedOrch aims to enhance the multi-agent collaborative performance of medical VQA tasks. The proposed system introduces three key agent categories: expert agents, a mediator agent, and a judge agent, as shown in Figure~\ref{fig:arch}. Expert agents are responsible for answering visual-language questions. The mediator agent is our key design, which synthesizes opinions, resolves conflicts, and facilitates discussions. The judge agent conducts the summarization and makes the final decision. The mediator agent and judge agent are driven by specific prompts. See Appendix Section Prompts for more details.

% A key design reflects on the role of mediator agent that synthesizes opinions, resolves conflicts, and facilitates discussions. A set of expert agents is responsible for answering visual-language questions and the judge agent conducts the summarization and makes the final prediction.

\subsection{Mediator-Guided Multi-Agent Collaboration}
\subsubsection{I. Initial VQA Stage.}
Upon receiving a medical question, multiple expert agents generate preliminary responses in parallel. Each agent then independently reasons through the question and forms a response, contributing to a diverse set of initial opinions. Initial opinions consist of the question analysis and answer options.

\subsubsection{II. Multi-Agent Collaboration.}
The mediator agent plays a central role in guiding the communication among expert agents. Its core responsibilities include analyzing the viewpoints from multiple expert agents, identifying opinion conflicts, and driving additional rounds of necessary discussions. 

\textbf{Analysis of expert agent's answer.} The mediator agent first systematically compares the initial responses from all expert agents. It identifies areas of consensus and disagreement, and flags responses that exhibit logical inconsistencies or semantic ambiguities, thereby laying the foundation for targeted follow-up discussion.

\textbf{Decision on whether additional discussion is needed.} Based on the comparative analysis, the mediator agent determines whether an additional dialogue with expert agents is required. If no significant discrepancies are identified, the decision is marked as ``No". Conversely, if critical issues arise or unclear reasoning are detected, it initiates a focused dialogue with the relevant expert agents to resolve these concerns, and the decision is recorded as ``Yes".

\textbf{Socratic questioning.} Once it is determined that further discussion is needed, the mediator agent engages in structured Socratic dialogues with the expert agents. The dialogues aim to promote information clarification, disaggregate reasoning, and resolve core disagreements. Each round of dialogue follows a well-defined format, including: \textbf{\textit{Opening the statement.}} The mediator agent begins by asserting itself prompted as ``the final authority on medical decisions" responsible for summarizing, reviewing, and synthesizing the opinions of all agents who act as experts. \textit{\textbf{Summary of other agents.}} To facilitate information sharing and encourage collaboration, the mediator agent summarizes the viewpoints of other agents and highlights points of disagreement with the current agent's position. \textit{\textbf{Review of the current agent.}} The mediator agent reviews the core conclusions of the current agent, e.g., ``You mentioned ..." and presents the reasoning behind their argument. \textit{\textbf{Analysis and questioning}}. Finally, the mediator agent raises in-depth questions targeting the main points of contention, ensuring the inquiry remains focused on substantial conflicts rather than vague or broad questions.

% \textbf{Automatic interactive instruction extraction.} To automate the interaction between the VLM and the mediator agent, the next step involves extracting interactive instructions from the mediator’s plain-text output to be fed back into the expert agent. To achieve this, a regular expression is employed to extract interactive instructions, including agent IDs and the corresponding question. This key mechanism ensures a consistent and stable communication across agents.

\textbf{Feedback-based answering.} Upon receiving the instruction from the mediator agent, the relevant expert agents reconstruct their initial answers based on the question. This becomes an evolutionary process that involves reinterpreting the original image content, correcting logical errors, and clarifying inconsistencies. Through this refinement, the overall quality, coherence, and explanatory depth of the final responses are evidently enhanced.

\subsubsection{III. Final Judgment.}
This stage is executed by the judge agent that possesses specialized auxiliary diagnostic capabilities. Its major responsibility is to integrate the feedback and make a scientifically sound decision. The judge agent is capable of handling complex opinion conflicts, extracting key arguments, and identifying areas of consensus and dispute. The judge agent serves as a final reviewer in the medical image assessment, synthesizing expert opinions and supporting the decision-making. In detail, it analyzes the complete dialogue between the mediator agent and the VLMs agent and organizes their core conclusions by extracting areas of consensus and disagreement from these agents. Based on this analysis, the judge agent combines the mediator agent’s focus to infer the conclusion and make the final diagnostic judgment.

\begin{table*}[!t]
\centering
\begin{tabular}{lccccc}
\toprule
Method                     & ~~~~~SLAKE~~~~~                                  & ~~~~~VQA-RAD~~~~~                                & ~~~~~PathVQA~~~~~                                & ~~~~~PMC-VQA~~~~~                                & ~~~~~Average~~~~~                                \\
\midrule
Qwen2.5-VL-7B        & 66.35                                  & 68.92                                  & 61.72                                  & 46.20                                  & 60.80                                  \\
HuatuoGPT-Vision-7B  & 70.19                                  & 74.10                                  & 60.81                                  & 46.15                                  & 62.81                                  \\
LLaVA-Med-7B         & 53.37                                  & 58.96                                  & 59.51                                  & 31.55                                  & 50.85                                  \\
Qwen2.5-VL-32B           & 65.87                                  & 74.10                                  & 64.85                                  & 48.75                                  & 63.39                                  \\
HuatuoGPT-Vision-34B & 68.99                                  & 75.70                                  & 62.61                                  & 50.05                                  & 64.34                                  \\
LLaVA-Next-34B            & 57.93                                  & 59.76                                  & 53.76                                  & 38.90                                  & 52.59                                  \\
\midrule
Judgment                & 69.71                                  & 69.72                                  & 60.63                                  & 48.75                                  & 62.20                                  \\
Voting                & 65.63                                  & 72.51                                  & 63.14                                  & 46.00                                  & 61.82                                  \\
MoA                  & 72.84                                  & 73.31                                  & 65.03                                  & 42.15                                  & 63.33                                  \\
Ours$^\blacklozenge$                & 68.75 & 74.10 & 64.26 & 46.40 & 63.38 \\
Ours$^\blacklozenge$ ↑Max/Min Gap & +15.38/--1.44          & +15.14/+0.00           & +4.75/+2.54             & +14.85/+0.20            & +12.53/+0.57            \\
Ours$^\clubsuit$                 & 73.32          & \textbf{79.28} & 64.82          & \textbf{50.50} & \textbf{66.98} \\
Ours$^\clubsuit$ ↑Max/Min Gap & +15.39/+4.33         & +19.52/+3.58            & +11.06/-0.03            & +11.60/+0.45            & +14.39/+2.64            \\
Ours$^\spadesuit$                 & \textbf{76.20} & 76.10          & \textbf{65.14} & 50.40          & 66.96         \\
Ours$^\spadesuit$ ↑Max/Min Gap & +10.33/+6.01          & +2.00/+0.40            & +4.33/+0.29             & +4.25/+0.35             & +4.15/+2.62             \\
\bottomrule
\end{tabular}
\caption{Results on four benchmarks. ``↑Max/Min Gap" indicates the performance gain of our  multi-agent method over the worst-performing and best-performing single agent for each dataset. The symbol $\blacklozenge$ represents the combination of three 7B models. The symbol $\clubsuit$ represents the combination of three $\sim$32B models. The symbol $\spadesuit$ represents the combination of HuatuoGPT-Vision-7B, Qwen2.5-VL-32B, and HuatuoGPT-Vision-34B.}
\label{four-benchmarks}
\end{table*}

\section{Experiments and Results}
\subsection{Experiment Settings}
\textbf{Benchmarks.} We use five widely-adopted medical VQA benchmarks, including VQA-RAD \cite{vqa-rad}, SLAKE \cite{slake}, PathVQA \cite{pathvqa}, PMC-VQA \cite{pmcvqa}, and a comprehensive benchmark OmniMedVQA \cite{omnimedvqa}. VQA-RAD centers on radiological images, and SLAKE encompasses a diverse range of medical image knowledge-intensive QA tasks. PathVQA focuses on tissue-level pathology-related images and questions. PMC-VQA is derived from PubMed Central, comprising 2,000 human-annotated QA pairs. For OmniMedVQA, we employ its publicly available subset including eight imaging modalities: computed tomography (CT), dermatological images (Der), fundus photography (FP), magnetic resonance imaging (MR), microscopic images (Mic), optical coherence tomography (OCT), ultrasound (US), and X-ray images. Detailed dataset statistics are provided in the Appendix Section: Benchmark Statistics.

\textbf{Details.} We systematically evaluate the performance of the proposed approach against multiple baseline methods in various settings. The baseline single-agent models included Qwen2.5-VL-7B \cite{qwen2.5vl}, HuatuoGPT-Vision-7B \cite{huatuogptvision}, LLaVA-Med-7B \cite{llavamed}, Qwen2.5-VL-32B \cite{qwen2.5vl}, HuatuoGPT-Vision-34B \cite{huatuogptvision}, and LLaVA-Next-34B \cite{llava}, with LLaVA-Med and HuatuoGPT-Vision being specialized medical VLMs. In addition, three major multi-agent strategies were compared: judgment (our baseline), voting \cite{voting}, and MoA \cite{moa}. The experiments utilize three agent configurations and tested three distinct combinations: (1) a combination of three 7B models, including Qwen2.5-VL-7B, HuatuoGPT-Vision-7B, LLaVA-Med-7B, denoted symbol $\blacklozenge$, (2) a combination of three $\sim$ 32B models, including Qwen2.5-VL-32B, HuatuoGPT-Vision-34B, and LLaVA-Next-34B, denoted symbol $\clubsuit$,  and (3) a combination of HuatuoGPT-Vision-7B, Qwen2.5-VL-32B, and HuatuoGPT-Vision-34B, denoted symbol $\spadesuit$. For the traditional medical VQA benchmarks, judgment, voting, and MoA strategies all use three models of approximately 32B. For the OmniMedVQA dataset, the multi-agent methods used three 7B models. By default, both the mediator agent and judge agent are set to Qwen2.5-32B \cite{qwen2.5}. Furthermore, we compare our approach with existing state-of-the-art (SOTA) multi-agent methods, including Reconcile \cite{reconcile}, AutoGen \cite{autogen}, DyLAN \cite{DyLAN}, and MedAgents \cite{medagents}, with comparative results sourced from the MDAgents \cite{mdagents}. Since the MDAgents method is in the few-shot setting while we are conducting zero-shot, we do not compare with it. All open-source models are executed on a server equipped with 8 NVIDIA A100 GPUs, using the LMdeploy \cite{lmdeploy} and vLLM \cite{vllm} frameworks for acceleration. In the experiment of ``Choice on LLM for mediator agents", the API call for GPT-4o-mini incurs a cost of \$2.02.

\textbf{Evaluation.} To ensure the consistency across agents when evaluating VLM performance on medical VQA tasks, we implement several standard processing steps. First, single-agent models are prompted using a unified instruction: ``provide the answer". Notably, we avoid imposing a rigid response format, such as requiring answers to be framed within predefined labels (e.g., ``A" or ``B"). This design allows each model to respond in its natural output style, thereby preserving the integrity of its reasoning process for subsequent analysis and integration by the mediator agent. Second, to extract structured answers from these unstructured outputs, we employ LLaMA-3.1-Instruct-8B \cite{llama3} as a parser. The parser standardizes model responses into an \verb|<answer> option </answer>| format, thereby improving the consistency with the downstream processing. At the final answer selection stage, inspired by the LVLM-eHub framework \cite{Lvlm-ehub}, we apply a semantic similarity matching between the parsed answer and the original candidate options. The option with the highest similarity score is selected as the final model's prediction. Accuracy is used as the evaluation metric in all experiments.

\subsection{Experimental Results}
\textbf{Results on the widely-adopted medical VQA benchmarks.} We systematically evaluate the MedOrch on four medical VQA benchmarks, including VQA-RAD \cite{vqa-rad}, SLAKE \cite{slake}, PathVQA \cite{pathvqa}, and PMC-VQA \cite{vqa-rad}. We compare our method against both single-agent and multi-agent methods. From Table \ref{four-benchmarks}, in the combination of small-scale models, ours$^\blacklozenge$ (comprising three 7B models) achieves an average accuracy of 63.38\%, outperforming all individual 7B models and multi-agent approaches. This demonstrates the effectiveness of inter-model collaboration even on a relatively modest scale.

On the larger scale, ours$^\clubsuit$ (comprising three $\sim$32B models) achieves the highest overall performance, with an average accuracy of 66.98\%, attaining the leading scores on VQA-RAD (79.28\%) and PMC-VQA (50.50\%). Compared to the best-performing single 32B model, this configuration yielded a performance improvement ranging from +0.03\% to +19.52\%, highlighting both the upper-bound performance potential of our multi-agent system across diverse medical VQA scenarios. LLaVA-Next-34B underperforms among the 32B models, achieving only 38.90\% accuracy on PMC-VQA, revealing its limited domain transferability in medical contexts. Notably, our multi-agent system maintains strong performance even when incorporating this weaker model, demonstrating resilience to the underperforming agent. As our baseline, the judgment method does not outperform the HuatuoGPT-Vision-34B in average performance. When integrating with our multi-agent collaboration framework, we find that the performance improves substantially (an average gain of 4.78\%). Other multi-agent methods exhibit higher sensitivity and generally fail to outperform the strongest individual agents. For instance, MoA exhibits strong performance on the PathVQA dataset but experiences a notable decline on the more challenging PMC-VQA dataset.

To further investigate the effect of model robustness, we construct a hybrid configuration ours$^\spadesuit$ by replacing LLaVA-Next-34B with the HuatuoGPT-Vision-7B model. This multi-agent framework achieved its best performance on SLAKE (76.20\%) and PathVQA (65.14\%), with an overall average accuracy of 66.96\%. These results underscore the collaboration value of diverse models within multi-agent systems.

We also report the performance gains relative to the weakest and best individual models (↑Max/Min Gap). Overall, our performance improvement is consistent across datasets given the design of our multi-agent workflow. For instance, ours$^\clubsuit$ outperforms the weakest model by +15.39\% on SLAKE and surpasses the best model by +4.33\%. This confirms that the proposed system can mitigate the limitations of weaker agents and amplify the strength of stronger ones via the proposed multi-agent collaboration system. A detailed case is shown in Appendix Section: Case Study.

\begin{table*}[!t]
\resizebox{\textwidth}{!}{
\begin{tabular}{lcccccccc}
\toprule
Method                  & CT                                     & Der                                    & FP                                     & MR                                     & Mic                                    & OCT                           & X-Ray                                  & US                            \\
                    \midrule
Qwen2.5-VL-7B       & 55.58                                  & 66.52                                  & 73.01                                  & 56.50                                  & 64.72                                  & 59.08                         & 73.94                                  & 37.08                         \\
HuatuoGPT-Vision-7B & 46.34                                  & 65.49                                  & 73.42                                  & 63.13                                  & 68.50                                  & \textbf{80.46}                & 77.40                                  & \textbf{76.41}                \\
LLaVA-Med-7B        & 52.38                                  & 56.67                                  & 47.67                                  & 49.20                                  & 47.75                                  & 55.29                         & 58.25                                  & 46.37                         \\
\midrule
Judgment               & 55.34                                  & 66.90                                  & 72.47                                  & 61.29                                  & 68.70                                  & 70.00                         & 76.34                                  & 51.83                         \\
Voting                & 44.99                                  & 64.82                                  & 69.71                                  & 59.55                                  & 63.70                                  & 68.77                         & 74.43                                  & 52.67                         \\
MoA                 & 50.26          & 66.52          & 73.40          & 60.12          & 63.98          & 60.87          & 74.13          & 39.24          \\
Ours$^\blacklozenge$                 & \textbf{61.29} & \textbf{67.73} & \textbf{76.10} & \textbf{63.90} & \textbf{70.90} & 70.75 & \textbf{77.92} & 60.28 \\
Ours$^\blacklozenge$
\scriptsize{↑Max/Min Gap}
                    & \scriptsize{+14.95/+5.71}
                    & \scriptsize{+11.06/+1.21}
                    & \scriptsize{+28.43/+2.68}
                    & \scriptsize{+14.70/+0.77}
                    & \scriptsize{+23.15/+2.40}
                    & \scriptsize{+15.46/--9.71}
                    & \scriptsize{+19.67/+0.52}
                    & \scriptsize{+23.20/--16.13} \\
\bottomrule
\end{tabular}}
\caption{Results on OmniMedVQA benchmark with eight modalities.}
\label{OmniMedVQA}
\vspace{-0.1cm}
\end{table*}

\begin{figure*}[!t]
    \centering
    \begin{subfigure}{0.32\linewidth}
        \centering
        \includegraphics[width=\linewidth]{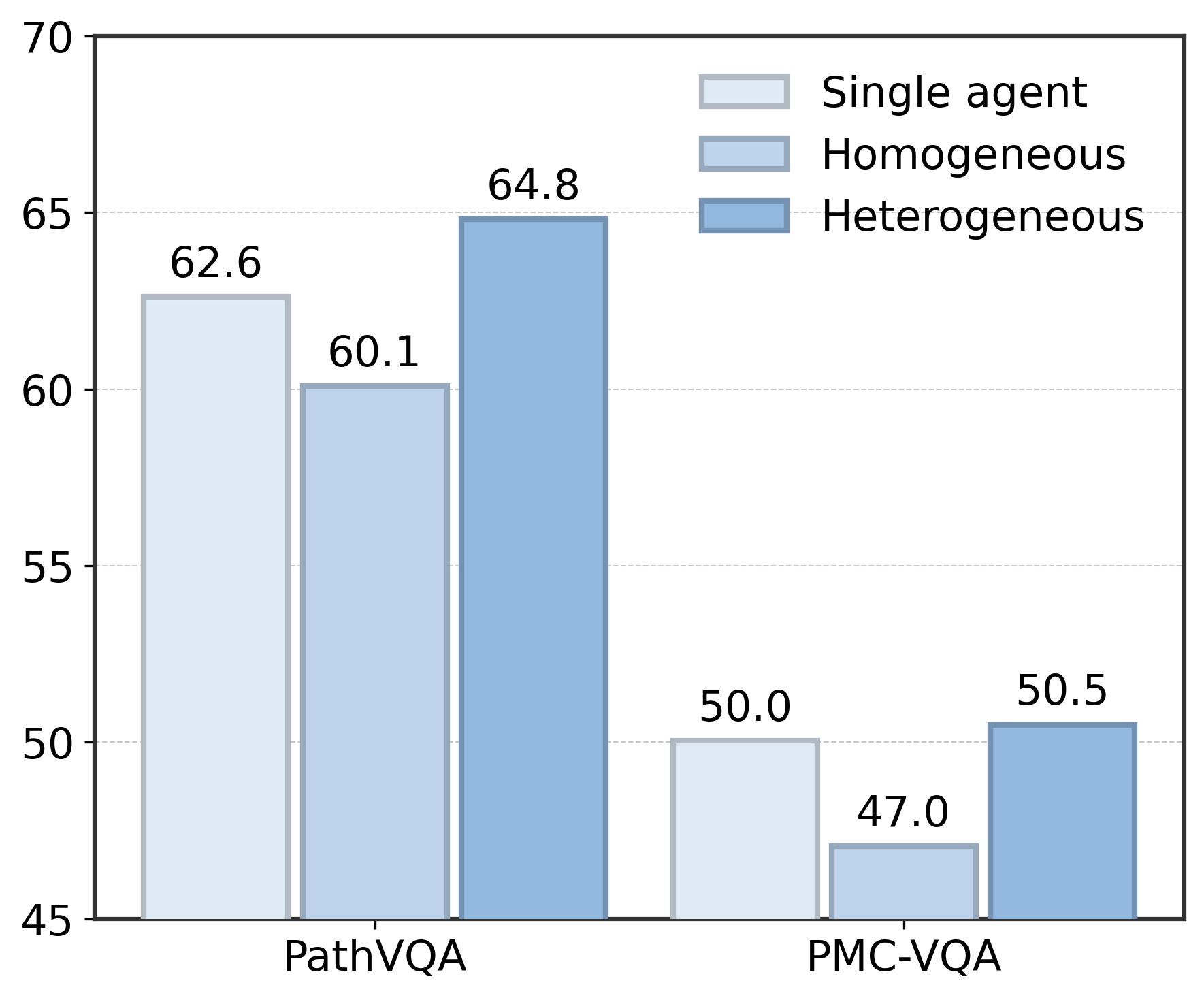}
        \caption{Results of using homogeneous agents.}
        \label{fig:homogeneous}
    \end{subfigure}
    \hfill
    \begin{subfigure}{0.32\linewidth}
        \centering
        \includegraphics[width=\linewidth]{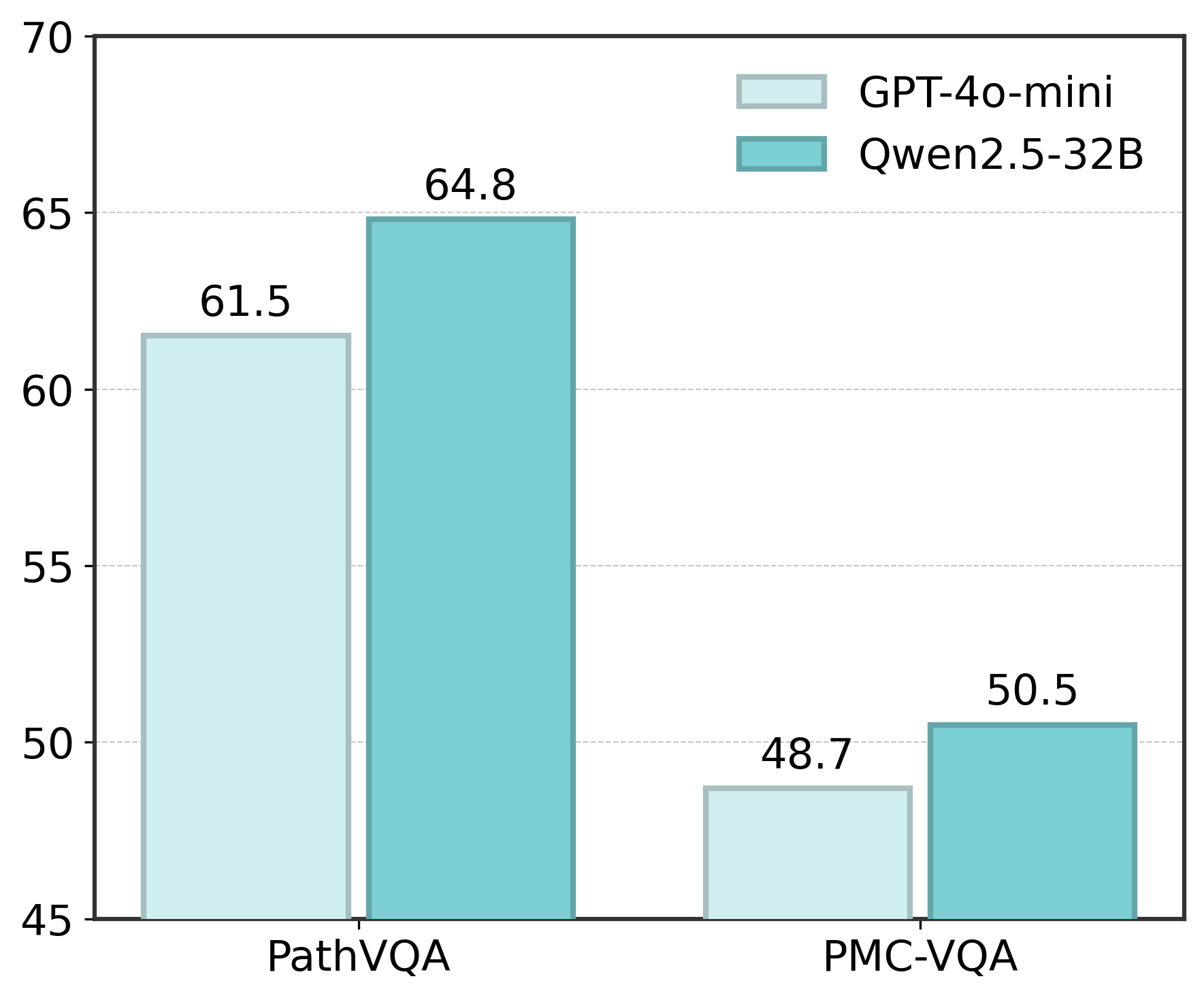}
        \caption{Results of different mediator agents.}
        \label{fig:mediator}
    \end{subfigure}
    \hfill
    \begin{subfigure}{0.32\linewidth}
        \centering
        \includegraphics[width=\linewidth]{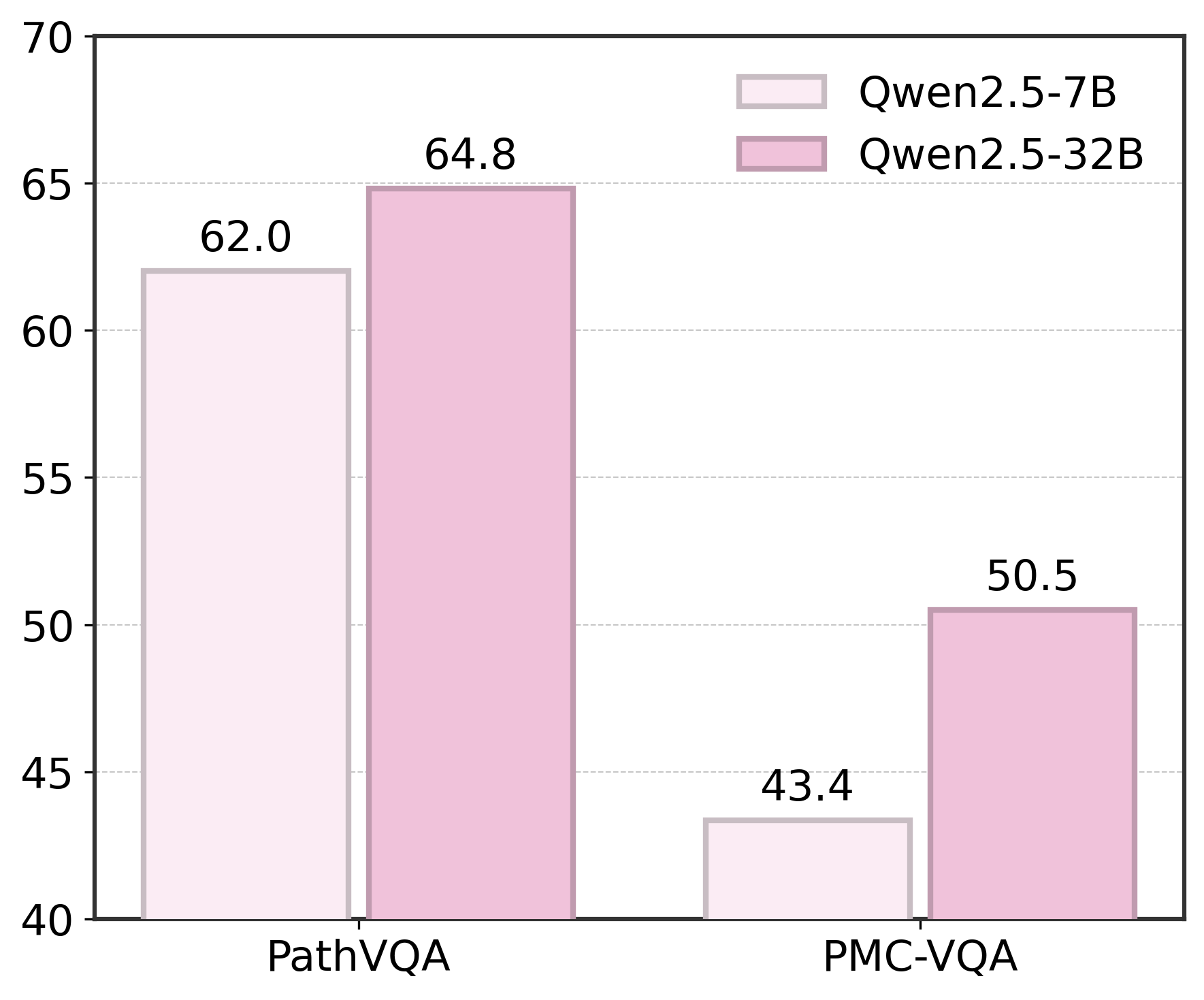}
        \caption{Results of different judge agents.}
        \label{fig:scale}
    \end{subfigure}
    \caption{Comparative results across different agent configurations. PathVQA and PMC-VQA are selected for these ablation studies as they are relatively more challenging and could provide presentative evaluations.}
    \label{fig:all_agents}
    \vspace{-0.3cm}
\end{figure*}

\textbf{Results on the OmniMedVQA.} We evaluate the proposed multi-agent framework on OmniMedVQA, a large-scale medical VQA dataset that spans eight imaging modalities. From Table \ref{OmniMedVQA}, ours$^\blacklozenge$, comprising multiple 7B models, exhibits strong performance among all evaluated approaches. It outperforms both multi-agent methods and the best-performing individual models. In particular, for the FP and Mic modalities, ours$^\blacklozenge$ achieved substantial accuracy improvements of +28.43\% and +23.15\%, respectively. These gains effectively compensated for individual model deficiencies in certain modalities, highlighting the framework’s capability to enhance modality-specific robustness. In the more challenging modalities, such as OCT and US, individual models like Qwen2.5-VL-7B and LLaVA-Med-7B exhibit relatively poor performance. Ours$^\blacklozenge$ does not outperform the strong single models yet surpasses other multi-agent methods, delivering notable improvements over the weakest agents with gains of +15.46\% and +23.20\%, respectively. These results demonstrate MedOrch's strong ability to suppress the negative influence of underperformed agents.

\begin{figure*}[!t]
    \centering
    \begin{minipage}[t]{0.48\linewidth}
        \centering
        \includegraphics[width=\linewidth]{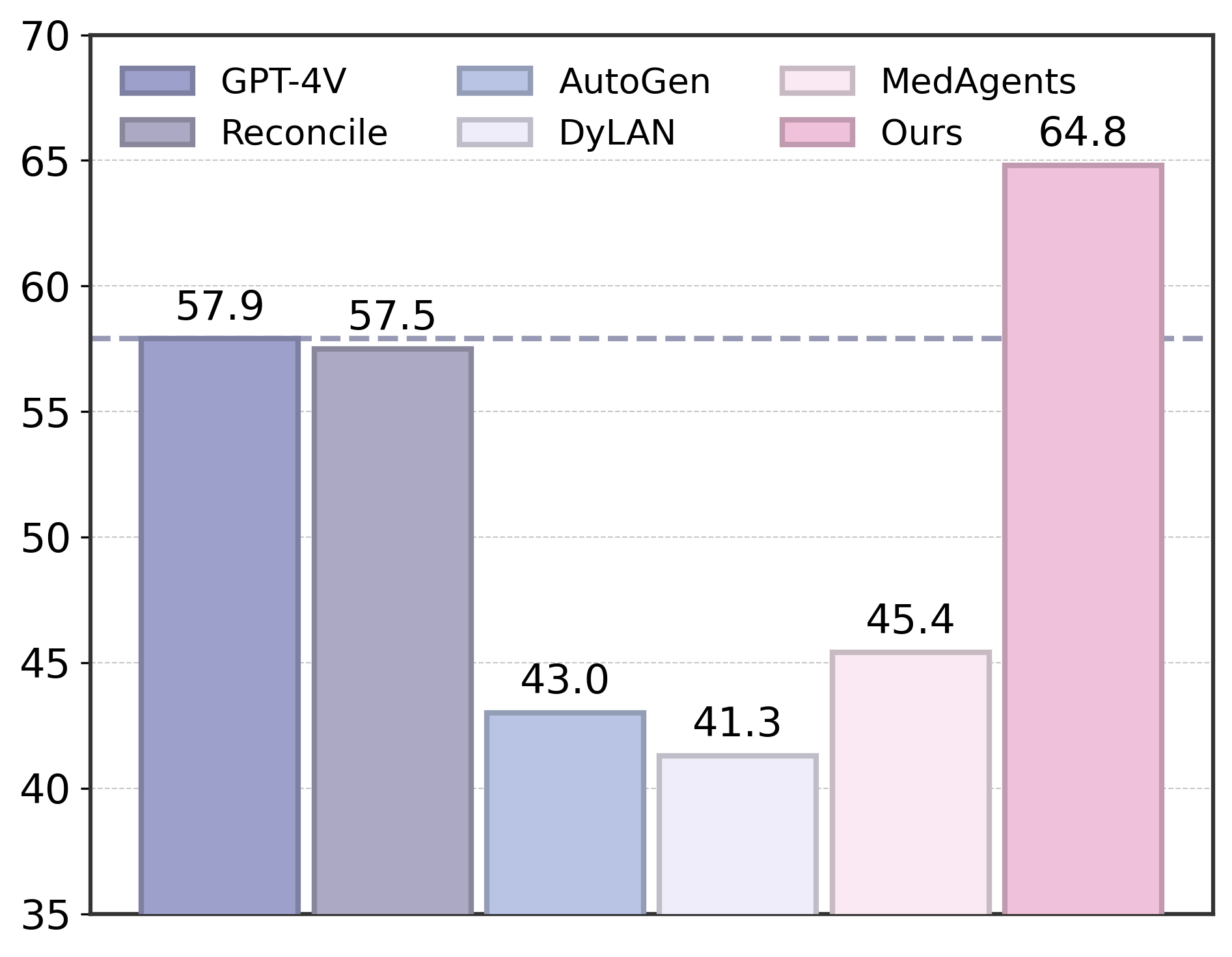}
        \caption{Comparison with GPT-4V and other multi-agent methods based on GPT-4V on PathVQA dataset.}
        \label{fig:method-comparison}
    \end{minipage}%
    \hfill
    \begin{minipage}[t]{0.48\linewidth}
        \centering
        % \raisebox{-2mm}{  
            \includegraphics[width=\linewidth]{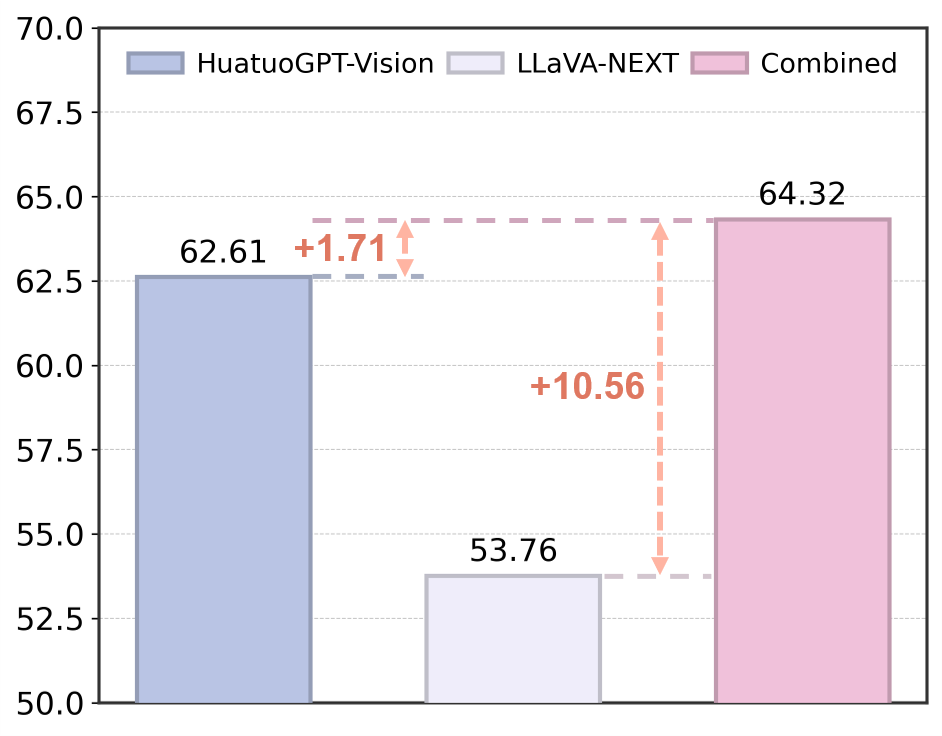}
        % }
        \caption{Evidence of collaborative synergy in MedOrch on PathVQA dataset.}
        \label{fig:two-agents}
    \end{minipage}
    \vspace{-0.4cm}
\end{figure*}

\textbf{Comparison with GPT-4V based multi-agent methods.} We further compare MedOrch, fully based on open-source models, against both GPT-4V and state-of-the-art multi-agent systems based on GPT-4V. As seen in Figure \ref{fig:method-comparison}, our method achieves an average accuracy of 64.8\%, substantially exceeding GPT-4V (57.9\%) and other multi-agent counterparts on the PathVQA dataset. In particular, we find that multi-agent methods \cite{autogen,DyLAN,medagents} originally designed for language tasks struggle to perform well in multimodal scenarios. Their performance evidently falls far behind the results achieved by using GPT-4V alone. In contrast, our multi-agent method demonstrates greater adaptability, underscoring the potential of an open-source compatible framework in medical multi-agent intelligence.

\textbf{Comparison with homogeneous multi-agent system.} To evaluate the performance of the multi-agent framework under a homogeneous model configuration, we construct multiple agents using the same base model of HuatuoGPT-Vision-34B. Specifically, we simulate three expert agents by running three independent inference trials of the same model, each with a temperature setting of 0.6 as seen in Figure \ref{fig:all_agents} (a). Under such a setting, multi-agent framework is unable to leverage the diversity across different models and thus we discourage the use of identical models. In contrast, we construct a heterogeneous multi-agent system composed of three different models towards performance gains. Notably, the two additional models, Qwen2.5-VL-32B and LLaVA-Next-34B, are weaker than HuatuoGPT-Vision-34B, as shown in Table~\ref{four-benchmarks}. This shows that agents with complementary perspectives can collaborate more effectively, even when some individual agents are less capable. Thus, a heterogeneous multi-agent system combining different VLMs is recommended.

\textbf{Choice on LLM for mediator agents.} We compare difference choices on LLM for mediator agents, including GPT-4o-mini (whose API fee is affordable) and Qwen2.5-32B. The results of PathVQA and PMC-VQA datasets are shown in Figure \ref{fig:all_agents} (b). Qwen2.5-32B achieves relatively high performance, with the accuracy of 64.8\% and 50.5\% in respective dataset. GPT-4o-mini shows a significant drop in performance, resulting in a lower overall accuracy. Note that we have explored the feasibility of using smaller models by evaluating a 7B-scale LLM as the mediator. However, we find that such small-scale models struggle to follow complex instructions and frequently fail to complete the tasks. Overall, these results highlight the value of using open-source LLMs with a proper scale for the mediator agent role in the multi-agent VQA system.

% \begin{figure}
%     \centering
%     \includegraphics[width=\linewidth]{Figures/scale_highlight_average.pdf}
%     \caption{Results of different scales of judge agents on four VQA benchmarks.}
%     \label{fig:scale}
% \end{figure}

% \begin{figure*}[!t]
% \centering
%   \begin{minipage}[]{0.44\textwidth}
%     \centering
%     \includegraphics[width=\textwidth]{Figures/number_highlighted_average.pdf}
%     \caption{Comparison with different number of expert agents on four VQA benchmarks.}
%     \label{fig:number}
%   \end{minipage}
%   \hspace{0.05\textwidth}
%   \begin{minipage}[]{0.44\textwidth}
%     \centering
%     \includegraphics[width=\textwidth]{Figures/scale_highlight_average.pdf}
%     \caption{Results of different scales of judge agents on four VQA benchmarks.}
%     \label{fig:scale}
%   \end{minipage}
% \vspace{-0.3cm}
% \end{figure*}

\textbf{Choice on LLM for judge agents.} To investigate the impact of judge agent on the performance, we compare the variants of the Qwen-2.5 series with 7B and 32B parameters, as summarized in Figure \ref{fig:all_agents} (c). The small-scale model, Qwen-2.5-7B, exhibits suboptimal performance, achieving the accuracy of 62.0\% on PathVQA and only 43.35\% on PMC-VQA. It indicates small-scale model has limited capacity for handling complex medical dialogues. Scaling up to 32B results in a substantial performance gain, achieving an accuracy of 64.8\% on PathVQA and 50.50\% on PMC-VQA. These findings underscore the effectiveness of models in semantic understanding and information integration. Qwen2.5-32B model offers a more favorable configuration for both mediator and judge agents, making it well-suited for deployment in practical applications.

% Our proposed multi-agent method demonstrate a clear advantage, even under homogeneous configurations, achieving an average accuracy of 63.41\%. This approach outperformed all single-agent baselines, with notable improvements on SLAKE (71.39\%) and PathVQA (60.69\%). These findings highlight that agent collaboration enhances reasoning capabilities and unlocks the representational potential of models, even when the underlying architecture is consistent across agents.

% \begin{figure}
%     \centering
%     \includegraphics[width=0.9\linewidth]{Figures/sota_grouped_bar.pdf}
%     \caption{Comparison with GPT-4V and other multi-agent methods based on GPT-4V.}
%     \label{fig:others}
% \end{figure}

% \begin{figure}
%     \centering
%     \includegraphics[width=0.9\linewidth]{Figures/merged_bar_chart.pdf}
%     \caption{Results of using homogeneous agents based on HuatuoGPT-Vision-34B}
%     \label{fig:homogeneous}
% \end{figure}

\textbf{``Conformity effect" or collaborative synergy?} We investigate whether the performance improvements observed during the collaboration stage are genuinely due to effective cooperation, or merely a result of the ``conformity effect". ``The conformity effect" refers to the tendency of individuals within a group to align with the majority opinion during the collaborative discussion.

To test this hypothesis, we introduce a two-agent configuration using HuatuoGPT-Vision-34B and LLaVA-Next-34B. In this two-agent setting, the possibility of ``minority yielding to majority" is strictly eliminated, making it a suitable setup to assess the effect of true collaborative synergy. Notably, we observe that performance improves even in this restricted two-agent setup, where the accuracy increases from 62.61\% (HuatuoGPT-Vision-34B alone) to 64.32\% with collaboration. This is particularly compelling given that LLaVA-Next-34B alone achieves only 53.76\% that is lower than HuatuoGPT-Vision-34B. When combined under our proposed framework, the result demonstrates an effect of multi-agent collaboration. These findings strongly suggest that the observed performance gains are not simply the result of conformity during group decision-making. Instead, they point to genuine complementarity and effective collaboration between heterogeneous agents.

\subsection{Discussion and Limitation}
% To construct an effective multimodal multi-agent system, we introduce LLM as a mediator, employing a Socratic questioning strategy to guide VLMs in reflecting on and refining responses. Experiments across multiple multimodal datasets demonstrate consistent performance gains over the best-performing single-agent models, with robustness maintained across various agent configurations and model scales.

% From the experiments, our study supports a heterogeneous agent system combining both general and domain-specific VLMs, as opposed to the mixture of agents (MoA) \cite{moa} on language tasks alone. 

Designing a collaborative system to handle multimodal clinical data is challenging given the complex healthcare workflow \cite{tu2024towards,acosta2022multimodal}. We explore the multi-agent collaboration from a collaboration perspective and measure its communication stability. In particular, our approach consistently outperformed the strongest single agent across diverse setups, including expert agent scale, composition, and mediator type. Towards a stable coordination mechanism, our approach leverages structured outputs generated by the mediator LLM and employs an automated regular-expression-based parsing to ensure a high coordination reliability. This setting strongly differs from AFlow \cite{aflow} that relies on search-based interactions yet suffers from its instability. Crucially, our findings reveal that heterogeneous multi-agent systems are particularly well-suited for leveraging open-source models in clinical applications.

Multi-agent frameworks inevitably introduce longer inference times compared to single-agent systems, primarily due to coordination overhead \cite{AgentSurvey2}. In our design, the main source of increased inference time stems from the collaborative stage. To reduce this, the mediator agent queries each expert agent only once without repeating the interaction process. In addition, we generate all questions of the mediator agent in a single forward pass and run multiple expert agents in parallel to handle responses. This yields an overall inference time approximately twice that of a single-agent system. We view this as a reasonable trade-off given the performance gains and believe the gap will further narrow with more optimizations of inference systems \cite{vllm, lmdeploy}.

Our approach relies entirely on the agents’ internal capabilities without the retrieval-augmented generation \cite{rag} and tools \cite{AgentSurvey2}. Incorporating real-world clinical data or medical knowledge can further enhance the system's performance in structured clinical workflows. Using specialized tools, such as imaging diagnostics and report generation, could advance the multi-agent collaboration and interpretable decision-making in the healthcare system.

% Although simulating differential agent roles via prompt engineering on GPT is possible, our results show limited effectiveness on multimodal decision-making. 
\section{Conclusion}
We propose MedOrch as a collaborative framework and verify that open-source multimodal agents can develop effective collaboration strategies for clinical decision-making. MedOrch builds upon the joint strength of open-source expert agents under the guidance of an LLM mediator agent. MedOrch enhances collaboration and addresses conflict resolution in complex medical conversations, such as error propagation and constrained inferential depth derived from individual VLMs. Comprehensive evaluations across five benchmark datasets demonstrate that our method attains strong performance without additional training. These findings underscore the potential of a reflective mediator-guided collaboration in enhancing medical multimodal intelligence.
\bibliography{main}

\begin{thebibliography}{47}
\providecommand{\natexlab}[1]{#1}

\bibitem[{Acosta et~al.(2022)Acosta, Falcone, Rajpurkar, and Topol}]{acosta2022multimodal}
Acosta, J.~N.; Falcone, G.~J.; Rajpurkar, P.; and Topol, E.~J. 2022.
\newblock Multimodal biomedical AI.
\newblock \emph{Nature medicine}, 28(9): 1773--1784.

\bibitem[{Bai et~al.(2025)Bai, Chen, Liu, Wang, Ge, Song, Dang, Wang, Wang, Tang et~al.}]{qwen2.5vl}
Bai, S.; Chen, K.; Liu, X.; Wang, J.; Ge, W.; Song, S.; Dang, K.; Wang, P.; Wang, S.; Tang, J.; et~al. 2025.
\newblock Qwen2. 5-vl technical report.
\newblock \emph{arXiv preprint arXiv:2502.13923}.

\bibitem[{Bansal et~al.(2024)Bansal, Israel, Zhao, Li, Nguyen, and Grover}]{MedMax}
Bansal, H.; Israel, D.; Zhao, S.; Li, S.; Nguyen, T.; and Grover, A. 2024.
\newblock MedMax: Mixed-Modal Instruction Tuning for Training Biomedical Assistants.
\newblock \emph{arXiv preprint arXiv:2412.12661}.

\bibitem[{Basit et~al.(2024)Basit, Hussain, Hanif, and Shafique}]{medaide}
Basit, A.; Hussain, K.; Hanif, M.~A.; and Shafique, M. 2024.
\newblock MedAide: Leveraging Large Language Models for On-Premise Medical Assistance on Edge Devices.
\newblock \emph{arXiv preprint arXiv:2403.00830}.

\bibitem[{Carey and Mullan(2004)}]{socratic2}
Carey, T.~A.; and Mullan, R.~J. 2004.
\newblock What is Socratic questioning?
\newblock \emph{Psychotherapy: theory, research, practice, training}, 41(3): 217.

\bibitem[{Chen et~al.(2024)Chen, Gui, Ouyang, Gao, Chen, Chen, Wang, Zhang, Cai, Ji et~al.}]{huatuogptvision}
Chen, J.; Gui, C.; Ouyang, R.; Gao, A.; Chen, S.; Chen, G.~H.; Wang, X.; Zhang, R.; Cai, Z.; Ji, K.; et~al. 2024.
\newblock Huatuogpt-vision, towards injecting medical visual knowledge into multimodal llms at scale.
\newblock \emph{arXiv preprint arXiv:2406.19280}.

\bibitem[{Chen, Saha, and Bansal(2023)}]{reconcile}
Chen, J. C.-Y.; Saha, S.; and Bansal, M. 2023.
\newblock Reconcile: Round-table conference improves reasoning via consensus among diverse llms.
\newblock \emph{arXiv preprint arXiv:2309.13007}.

\bibitem[{Chen et~al.(2025)Chen, Li, Yang, Wang, Dong, and Gao}]{MDTeamGPT}
Chen, K.; Li, X.; Yang, T.; Wang, H.; Dong, W.; and Gao, Y. 2025.
\newblock MDTeamGPT: A Self-Evolving LLM-based Multi-Agent Framework for Multi-Disciplinary Team Medical Consultation.
\newblock \emph{arXiv preprint arXiv:2503.13856}.

\bibitem[{Cheng et~al.(2024)Cheng, Zhang, Zhang, Meng, Hong, Li, Wang, Wang, Yin, Zhao et~al.}]{AgentSurvey2}
Cheng, Y.; Zhang, C.; Zhang, Z.; Meng, X.; Hong, S.; Li, W.; Wang, Z.; Wang, Z.; Yin, F.; Zhao, J.; et~al. 2024.
\newblock Exploring large language model based intelligent agents: Definitions, methods, and prospects.
\newblock \emph{arXiv preprint arXiv:2401.03428}.

\bibitem[{Contributors(2023)}]{lmdeploy}
Contributors, L. 2023.
\newblock LMDeploy: A Toolkit for Compressing, Deploying, and Serving LLM.
\newblock \url{https://github.com/InternLM/lmdeploy}.

\bibitem[{Du et~al.(2023)Du, Li, Torralba, Tenenbaum, and Mordatch}]{debate}
Du, Y.; Li, S.; Torralba, A.; Tenenbaum, J.~B.; and Mordatch, I. 2023.
\newblock Improving factuality and reasoning in language models through multiagent debate.
\newblock In \emph{Forty-first International Conference on Machine Learning}.

\bibitem[{Elder and Paul(1998)}]{Socratic}
Elder, L.; and Paul, R. 1998.
\newblock The role of Socratic questioning in thinking, teaching, and learning.
\newblock \emph{The Clearing House}, 71(5): 297--301.

\bibitem[{Grattafiori et~al.(2024)Grattafiori, Dubey, Jauhri, Pandey, Kadian, Al-Dahle, Letman, Mathur, Schelten, Vaughan et~al.}]{llama3}
Grattafiori, A.; Dubey, A.; Jauhri, A.; Pandey, A.; Kadian, A.; Al-Dahle, A.; Letman, A.; Mathur, A.; Schelten, A.; Vaughan, A.; et~al. 2024.
\newblock The llama 3 herd of models.
\newblock \emph{arXiv preprint arXiv:2407.21783}.

\bibitem[{He et~al.(2020)He, Zhang, Mou, Xing, and Xie}]{pathvqa}
He, X.; Zhang, Y.; Mou, L.; Xing, E.; and Xie, P. 2020.
\newblock Pathvqa: 30000+ questions for medical visual question answering.
\newblock \emph{arXiv preprint arXiv:2003.10286}.

\bibitem[{Hu et~al.(2024)Hu, Li, Lu, Shao, He, Qiao, and Luo}]{omnimedvqa}
Hu, Y.; Li, T.; Lu, Q.; Shao, W.; He, J.; Qiao, Y.; and Luo, P. 2024.
\newblock Omnimedvqa: A new large-scale comprehensive evaluation benchmark for medical lvlm.
\newblock In \emph{Proceedings of the IEEE/CVF Conference on Computer Vision and Pattern Recognition}, 22170--22183.

\bibitem[{Kim et~al.(2024)Kim, Park, Jeong, Chan, Xu, McDuff, Lee, Ghassemi, Breazeal, Park et~al.}]{mdagents}
Kim, Y.; Park, C.; Jeong, H.; Chan, Y.~S.; Xu, X.; McDuff, D.; Lee, H.; Ghassemi, M.; Breazeal, C.; Park, H.; et~al. 2024.
\newblock Mdagents: An adaptive collaboration of llms for medical decision-making.
\newblock \emph{Advances in Neural Information Processing Systems}, 37: 79410--79452.

\bibitem[{Kwon et~al.(2023)Kwon, Li, Zhuang, Sheng, Zheng, Yu, Gonzalez, Zhang, and Stoica}]{vllm}
Kwon, W.; Li, Z.; Zhuang, S.; Sheng, Y.; Zheng, L.; Yu, C.~H.; Gonzalez, J.~E.; Zhang, H.; and Stoica, I. 2023.
\newblock Efficient Memory Management for Large Language Model Serving with PagedAttention.
\newblock In \emph{Proceedings of the ACM SIGOPS 29th Symposium on Operating Systems Principles}.

\bibitem[{Lau et~al.(2018)Lau, Gayen, Ben~Abacha, and Demner-Fushman}]{vqa-rad}
Lau, J.~J.; Gayen, S.; Ben~Abacha, A.; and Demner-Fushman, D. 2018.
\newblock A dataset of clinically generated visual questions and answers about radiology images.
\newblock \emph{Scientific data}, 5(1): 1--10.

\bibitem[{Lewis et~al.(2020)Lewis, Perez, Piktus, Petroni, Karpukhin, Goyal, K{\"u}ttler, Lewis, Yih, Rockt{\"a}schel et~al.}]{rag}
Lewis, P.; Perez, E.; Piktus, A.; Petroni, F.; Karpukhin, V.; Goyal, N.; K{\"u}ttler, H.; Lewis, M.; Yih, W.-t.; Rockt{\"a}schel, T.; et~al. 2020.
\newblock Retrieval-augmented generation for knowledge-intensive nlp tasks.
\newblock \emph{Advances in neural information processing systems}, 33: 9459--9474.

\bibitem[{Li et~al.(2023{\natexlab{a}})Li, Wong, Zhang, Usuyama, Liu, Yang, Naumann, Poon, and Gao}]{llavamed}
Li, C.; Wong, C.; Zhang, S.; Usuyama, N.; Liu, H.; Yang, J.; Naumann, T.; Poon, H.; and Gao, J. 2023{\natexlab{a}}.
\newblock Llava-med: Training a large language-and-vision assistant for biomedicine in one day.
\newblock \emph{Advances in Neural Information Processing Systems}, 36: 28541--28564.

\bibitem[{Li et~al.(2023{\natexlab{b}})Li, Hammoud, Itani, Khizbullin, and Ghanem}]{camel}
Li, G.; Hammoud, H.; Itani, H.; Khizbullin, D.; and Ghanem, B. 2023{\natexlab{b}}.
\newblock Camel: Communicative agents for" mind" exploration of large language model society.
\newblock \emph{Advances in Neural Information Processing Systems}, 36: 51991--52008.

\bibitem[{Li et~al.(2024)Li, Lai, Li, Ren, Zhang, Kang, Wang, Li, Zhang, Ma et~al.}]{li}
Li, J.; Lai, Y.; Li, W.; Ren, J.; Zhang, M.; Kang, X.; Wang, S.; Li, P.; Zhang, Y.-Q.; Ma, W.; et~al. 2024.
\newblock Agent hospital: A simulacrum of hospital with evolvable medical agents.
\newblock \emph{arXiv preprint arXiv:2405.02957}.

\bibitem[{Liang et~al.(2023)Liang, He, Jiao, Wang, Wang, Wang, Yang, Shi, and Tu}]{MAD}
Liang, T.; He, Z.; Jiao, W.; Wang, X.; Wang, Y.; Wang, R.; Yang, Y.; Shi, S.; and Tu, Z. 2023.
\newblock Encouraging divergent thinking in large language models through multi-agent debate.
\newblock \emph{arXiv preprint arXiv:2305.19118}.

\bibitem[{Liu et~al.(2021)Liu, Zhan, Xu, Ma, Yang, and Wu}]{slake}
Liu, B.; Zhan, L.-M.; Xu, L.; Ma, L.; Yang, Y.; and Wu, X.-M. 2021.
\newblock Slake: A semantically-labeled knowledge-enhanced dataset for medical visual question answering.
\newblock In \emph{2021 IEEE 18th international symposium on biomedical imaging (ISBI)}, 1650--1654. IEEE.

\bibitem[{Liu et~al.(2023{\natexlab{a}})Liu, Li, Wu, and Lee}]{llava}
Liu, H.; Li, C.; Wu, Q.; and Lee, Y.~J. 2023{\natexlab{a}}.
\newblock Visual instruction tuning.
\newblock \emph{Advances in neural information processing systems}, 36: 34892--34916.

\bibitem[{Liu et~al.(2023{\natexlab{b}})Liu, Zhang, Li, Liu, and Yang}]{DyLAN}
Liu, Z.; Zhang, Y.; Li, P.; Liu, Y.; and Yang, D. 2023{\natexlab{b}}.
\newblock Dynamic llm-agent network: An llm-agent collaboration framework with agent team optimization.
\newblock \emph{arXiv preprint arXiv:2310.02170}.

\bibitem[{Long et~al.(2024)Long, Yen, Luu, Kawaguchi, Kan, and Chen}]{long}
Long, D.~X.; Yen, D.~N.; Luu, A.~T.; Kawaguchi, K.; Kan, M.-Y.; and Chen, N.~F. 2024.
\newblock Multi-expert Prompting Improves Reliability, Safety and Usefulness of Large Language Models.
\newblock In \emph{Proceedings of the 2024 Conference on Empirical Methods in Natural Language Processing}, 20370--20401.

\bibitem[{Luo et~al.(2025)Luo, Zhang, Yuan, Zhao, Yang, Gu, Wu, Chen, Qiao, Long et~al.}]{AgentSurvey1}
Luo, J.; Zhang, W.; Yuan, Y.; Zhao, Y.; Yang, J.; Gu, Y.; Wu, B.; Chen, B.; Qiao, Z.; Long, Q.; et~al. 2025.
\newblock Large Language Model Agent: A Survey on Methodology, Applications and Challenges.
\newblock \emph{arXiv preprint arXiv:2503.21460}.

\bibitem[{Ma et~al.(2024)Ma, Hu, Pu, Boyin, Ai, Liang, and Chen}]{cory}
Ma, H.; Hu, T.; Pu, Z.; Boyin, L.; Ai, X.; Liang, Y.; and Chen, M. 2024.
\newblock Coevolving with the other you: Fine-tuning llm with sequential cooperative multi-agent reinforcement learning.
\newblock \emph{Advances in Neural Information Processing Systems}, 37: 15497--15525.

\bibitem[{Pandey, Amod et~al.(2024)}]{pandey2024advancing}
Pandey, H.; Amod, A.; et~al. 2024.
\newblock Advancing Healthcare Automation: Multi-agent system for medical necessity justification.
\newblock \emph{arXiv preprint arXiv:2404.17977}.

\bibitem[{Singhal et~al.(2023)Singhal, Azizi, Tu, Mahdavi, Wei, Chung, Scales, Tanwani, Cole-Lewis, Pfohl et~al.}]{llmmedicine2}
Singhal, K.; Azizi, S.; Tu, T.; Mahdavi, S.~S.; Wei, J.; Chung, H.~W.; Scales, N.; Tanwani, A.; Cole-Lewis, H.; Pfohl, S.; et~al. 2023.
\newblock Large language models encode clinical knowledge.
\newblock \emph{Nature}, 620(7972): 172--180.

\bibitem[{Sun et~al.(2024)Sun, Zhang, Si, Zhu, Shui, Zhang, Li, Lyu, Lin, and Yang}]{pathgen}
Sun, Y.; Zhang, Y.; Si, Y.; Zhu, C.; Shui, Z.; Zhang, K.; Li, J.; Lyu, X.; Lin, T.; and Yang, L. 2024.
\newblock Pathgen-1.6 m: 1.6 million pathology image-text pairs generation through multi-agent collaboration.
\newblock \emph{arXiv preprint arXiv:2407.00203}.

\bibitem[{Tang et~al.(2023)Tang, Zou, Zhang, Li, Zhao, Zhang, Cohan, and Gerstein}]{medagents}
Tang, X.; Zou, A.; Zhang, Z.; Li, Z.; Zhao, Y.; Zhang, X.; Cohan, A.; and Gerstein, M. 2023.
\newblock Medagents: Large language models as collaborators for zero-shot medical reasoning.
\newblock \emph{arXiv preprint arXiv:2311.10537}.

\bibitem[{Thirunavukarasu et~al.(2023)Thirunavukarasu, Ting, Elangovan, Gutierrez, Tan, and Ting}]{llmmedicine1}
Thirunavukarasu, A.~J.; Ting, D. S.~J.; Elangovan, K.; Gutierrez, L.; Tan, T.~F.; and Ting, D. S.~W. 2023.
\newblock Large language models in medicine.
\newblock \emph{Nature medicine}, 29(8): 1930--1940.

\bibitem[{Tu et~al.(2024)Tu, Azizi, Driess, Schaekermann, Amin, Chang, Carroll, Lau, Tanno, Ktena et~al.}]{tu2024towards}
Tu, T.; Azizi, S.; Driess, D.; Schaekermann, M.; Amin, M.; Chang, P.-C.; Carroll, A.; Lau, C.; Tanno, R.; Ktena, I.; et~al. 2024.
\newblock Towards generalist biomedical AI.
\newblock \emph{Nejm Ai}, 1(3): AIoa2300138.

\bibitem[{Wang et~al.(2024)Wang, Wang, Athiwaratkun, Zhang, and Zou}]{moa}
Wang, J.; Wang, J.; Athiwaratkun, B.; Zhang, C.; and Zou, J. 2024.
\newblock Mixture-of-agents enhances large language model capabilities.
\newblock \emph{arXiv preprint arXiv:2406.04692}.

\bibitem[{Wang et~al.(2022)Wang, Wei, Schuurmans, Le, Chi, Narang, Chowdhery, and Zhou}]{voting}
Wang, X.; Wei, J.; Schuurmans, D.; Le, Q.; Chi, E.; Narang, S.; Chowdhery, A.; and Zhou, D. 2022.
\newblock Self-consistency improves chain of thought reasoning in language models.
\newblock \emph{arXiv preprint arXiv:2203.11171}.

\bibitem[{Wei et~al.(2024)Wei, Qiu, Yu, and Yuan}]{medco}
Wei, H.; Qiu, J.; Yu, H.; and Yuan, W. 2024.
\newblock Medco: Medical education copilots based on a multi-agent framework.
\newblock \emph{arXiv preprint arXiv:2408.12496}.

\bibitem[{Wu et~al.(2023{\natexlab{a}})Wu, Lei, Zheng, Zhao, Lin, Zhang, Zhou, Zhao, Zhang, Wang et~al.}]{gpt-4vmedical}
Wu, C.; Lei, J.; Zheng, Q.; Zhao, W.; Lin, W.; Zhang, X.; Zhou, X.; Zhao, Z.; Zhang, Y.; Wang, Y.; et~al. 2023{\natexlab{a}}.
\newblock Can gpt-4v (ision) serve medical applications? case studies on gpt-4v for multimodal medical diagnosis.
\newblock \emph{arXiv preprint arXiv:2310.09909}.

\bibitem[{Wu et~al.(2023{\natexlab{b}})Wu, Bansal, Zhang, Wu, Li, Zhu, Jiang, Zhang, Zhang, Liu et~al.}]{autogen}
Wu, Q.; Bansal, G.; Zhang, J.; Wu, Y.; Li, B.; Zhu, E.; Jiang, L.; Zhang, X.; Zhang, S.; Liu, J.; et~al. 2023{\natexlab{b}}.
\newblock Autogen: Enabling next-gen llm applications via multi-agent conversation.
\newblock \emph{arXiv preprint arXiv:2308.08155}.

\bibitem[{Xu et~al.(2024)Xu, Shao, Zhang, Gao, Liu, Lei, Meng, Huang, Qiao, and Luo}]{Lvlm-ehub}
Xu, P.; Shao, W.; Zhang, K.; Gao, P.; Liu, S.; Lei, M.; Meng, F.; Huang, S.; Qiao, Y.; and Luo, P. 2024.
\newblock Lvlm-ehub: A comprehensive evaluation benchmark for large vision-language models.
\newblock \emph{IEEE Transactions on Pattern Analysis and Machine Intelligence}.

\bibitem[{Yang et~al.(2024)Yang, Yang, Zhang, Hui, Zheng, Yu, Li, Liu, Huang, Wei et~al.}]{qwen2.5}
Yang, A.; Yang, B.; Zhang, B.; Hui, B.; Zheng, B.; Yu, B.; Li, C.; Liu, D.; Huang, F.; Wei, H.; et~al. 2024.
\newblock Qwen2. 5 technical report.
\newblock \emph{arXiv preprint arXiv:2412.15115}.

\bibitem[{Zhang et~al.(2024{\natexlab{a}})Zhang, Yang, Hu, Wang, Li, Sun, Zhang, Zhang, Liu, Zhu et~al.}]{proagent}
Zhang, C.; Yang, K.; Hu, S.; Wang, Z.; Li, G.; Sun, Y.; Zhang, C.; Zhang, Z.; Liu, A.; Zhu, S.-C.; et~al. 2024{\natexlab{a}}.
\newblock Proagent: building proactive cooperative agents with large language models.
\newblock In \emph{Proceedings of the AAAI Conference on Artificial Intelligence}, volume~38, 17591--17599.

\bibitem[{Zhang et~al.(2024{\natexlab{b}})Zhang, Xiang, Yu, Teng, Chen, Chen, Zhuge, Cheng, Hong, Wang et~al.}]{aflow}
Zhang, J.; Xiang, J.; Yu, Z.; Teng, F.; Chen, X.; Chen, J.; Zhuge, M.; Cheng, X.; Hong, S.; Wang, J.; et~al. 2024{\natexlab{b}}.
\newblock Aflow: Automating agentic workflow generation.
\newblock \emph{arXiv preprint arXiv:2410.10762}.

\bibitem[{Zhang et~al.(2023)Zhang, Wu, Zhao, Lin, Zhang, Wang, and Xie}]{pmcvqa}
Zhang, X.; Wu, C.; Zhao, Z.; Lin, W.; Zhang, Y.; Wang, Y.; and Xie, W. 2023.
\newblock Pmc-vqa: Visual instruction tuning for medical visual question answering.
\newblock \emph{arXiv preprint arXiv:2305.10415}.

\bibitem[{Zhang et~al.(2024{\natexlab{c}})Zhang, Gao, Tan, Zhou, Ding, Zhou, Zhang, and Wang}]{zhang2024data}
Zhang, Y.; Gao, J.; Tan, Z.; Zhou, L.; Ding, K.; Zhou, M.; Zhang, S.; and Wang, D. 2024{\natexlab{c}}.
\newblock Data-centric foundation models in computational healthcare: A survey.
\newblock \emph{arXiv preprint arXiv:2401.02458}.

\bibitem[{Zuo et~al.(2025)Zuo, Qu, Li, Chen, Zhu, Hua, Zhang, Ding, and Zhou}]{medxpertqa}
Zuo, Y.; Qu, S.; Li, Y.; Chen, Z.; Zhu, X.; Hua, E.; Zhang, K.; Ding, N.; and Zhou, B. 2025.
\newblock MedXpertQA: Benchmarking Expert-Level Medical Reasoning and Understanding.
\newblock \emph{arXiv preprint arXiv:2501.18362}.

\end{thebibliography}

\onecolumn

% \begin{tcolorbox}[width=\textwidth, colback=blue!10] % 明确指定单栏宽度
% xxxxxxxxxxxxxxxxxxxxxxxxxxxxxxxxxxxxxxxxxxx
% \end{tcolorbox}

\tcbset{
  colback=white,    
  colframe=black,   
  fonttitle=\bfseries,
  coltitle=black,
  boxrule=0.5mm,
  arc=2mm,
  left=2mm,
  right=2mm,
  top=1mm,
  bottom=1mm
}

\section*{Case Study}

\begin{tcolorbox}[title=A Case of Multi-agent collaboration,
  % coltitle=white,  
  enhanced,
  breakable,       
  colbacktitle=mybluex,      
  colframe=black,   
overlay first={
    \node[anchor=north east, inner sep=11pt,yshift=-21pt     ]
      at (frame.north east)
      {\includegraphics[width=3.5cm]{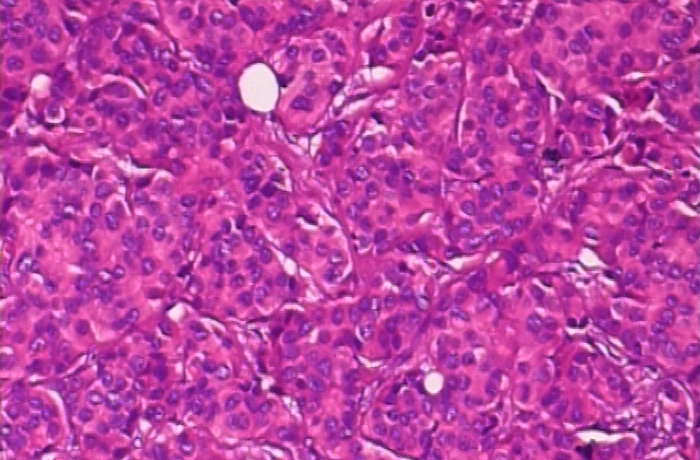}};
  }
 % watermark graphics=Figures/SOB_M_DC-14-9461-100-002.png,   % 图片路径
 %  watermark width=2cm,                   % 图片大小
 %  watermark at north east,               % 位置：右上
 %  watermark inner sep=2pt,               % 与边框的距离
 %  watermark opacity=1
]

\textbf{Question:} \\
What does this histological examination of breast tissue reveal? \\
(A): Malignant breast histopathology. \raisebox{-0.2\height}{\includegraphics[height=1em]{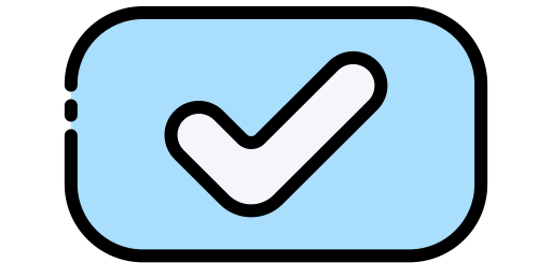}}\\
(B): Non-cancerous kidney histopathology. \\
(C): Normal brain histopathology. \\
(D): Inflammatory bowel disease histopathology. \\
Please provide the answer.\\

\textbf{Initial Response from Expert Agent 1 \raisebox{-0.1\height}{\includegraphics[height=1em]{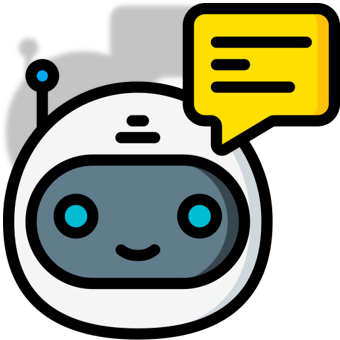}}:} \\
\textbf{\textit{\textcolor{myblue}{Option A: Malignant breast histopathology.}}} The image shows a dense proliferation of atypical cells with irregular nuclei and prominent nucleoli, which is characteristic of malignant tissue in the context provided (breast). This aligns well with features typically seen in malignancies such as invasive ductal carcinoma or other forms of breast cancer under microscopic examination using hematoxylin-eosin staining technique. Other options like non-cancerous kidney, normal brain, and inflammatory bowel disease would not display these specific cellular characteristics typical for their respective tissues when viewed microscopically stained similarly. \raisebox{-0.2\height}{\includegraphics[height=1em]{Figures/right.png}}\\

\textbf{Initial Response from Expert Agent 2 \raisebox{-0.1\height}{\includegraphics[height=1em]{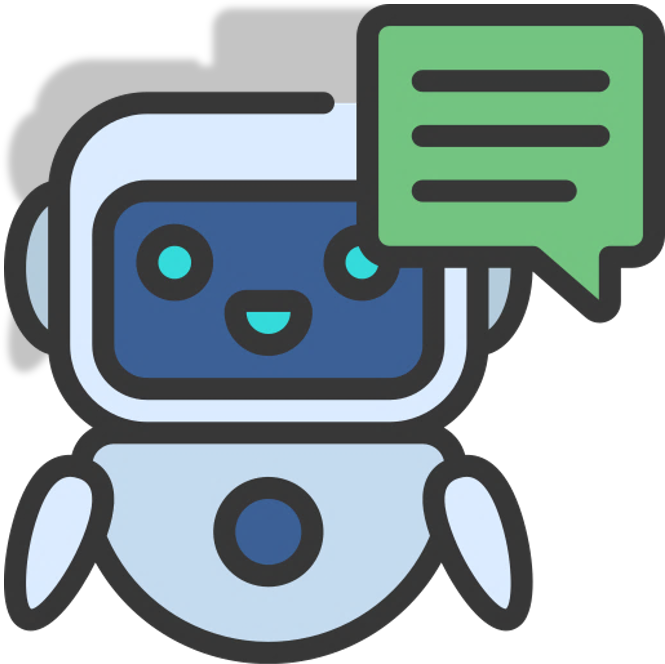}}:} \\
The image provided is a histological examination of breast tissue. The cells appear to be arranged in a pattern that is characteristic of breast tissue. There is no evidence of atypical or abnormal cell growth that would suggest malignancy. Given the options provided: Option A: Malignant breast histopathology - This is not supported by the image, as there are no features indicative of malignancy. Option B: Non-cancerous kidney histopathology - This is incorrect as the image is of breast tissue, not kidney tissue. Option C: Normal brain histopathology - This is incorrect as the image is of breast tissue, not brain tissue. Option D: Inflammatory bowel disease histopathology - This is incorrect as the image is of breast tissue, not bowel tissue. This is not supported by the image, as there are no features indicative of malignancy. \textcolor{myred}{\textbf{\textit{The image appears to show normal breast tissue.}}} \raisebox{-0.2\height}{\includegraphics[height=1em]{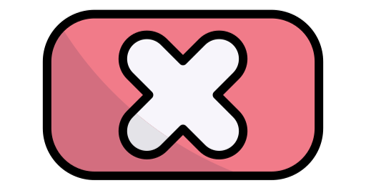}}\\

\textbf{Initial Response from Expert Agent 3 \raisebox{-0.1\height}{\includegraphics[height=1em]{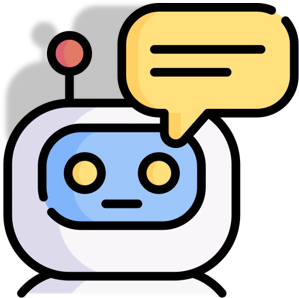}}:} \\
The histological examination of the breast tissue reveals \textcolor{myblue}{\textbf{\textit{malignant breast histopathology}}}, which indicates the presence of cancerous cells in the breast tissue. \raisebox{-0.2\height}{\includegraphics[height=1em]{Figures/right.png}}\\

\textbf{Question for Expert Agent 1 Posed by Mediator Agent \raisebox{-0.1\height}{\includegraphics[height=1em]{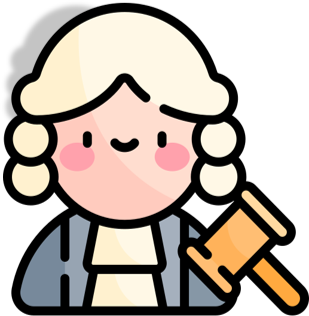}}:} \\
I am the final authority in medical decision-making, responsible for reviewing and synthesizing all opinions from diverse medical experts. Expert 2 and Expert 3 have differing opinions on the histological examination of the breast tissue. Expert 2 does not see any features indicative of malignancy and suggests the image shows normal breast tissue, while you mentioned that the image shows a dense proliferation of atypical cells with irregular nuclei and prominent nucleoli, characteristic of malignant tissue. \textit{Could you elaborate on the specific cellular features that led you to conclude the presence of malignancy, and how these features differ from those seen in normal breast tissue?}\\

\textbf{Response from Expert Agent 1  \raisebox{-0.1\height}{\includegraphics[height=1em]{Figures/a1.png}}:}\\
As an expert in breast histopathology, I can provide a detailed analysis of the cellular features observed in this image that indicate malignancy. The key findings include: 1) Dense proliferation: There is a dense and disorganized growth pattern of cells throughout the tissue section. 2) Atypical cell morphology: The tumor cells exhibit significant variations in size (pleomorphism), shape, and nuclear characteristics compared to normal breast epithelial cells. 3) Prominent nucleoli: Many nuclei appear enlarged with prominent nucleoli, which are indicative of increased protein synthesis required for rapid cell division. 4) Irregular nuclear contours: Some nuclei have irregular or indented borders instead of smooth outlines seen in benign cells. 5) Increased mitotic activity: An elevated number of dividing cells (mitoses) suggests uncontrolled cell proliferation typical of malignant tumors. 6) Loss of architectural organization: Unlike normal ductal structures found in non-cancerous breast tissue, these atypical cells lack any recognizable glandular patterns or organized architecture.  These cellular abnormalities collectively point towards a diagnosis of invasive carcinoma rather than normal breast parenchyma or inflammatory processes like those seen in inflammatory bowel disease. While some benign lesions may also display mild cytological changes, the degree of pleomorphism, high mitotic rate, loss of differentiation, and overall disorganization strongly suggest a malignant neoplastic process consistent with \textcolor{myblue}{\textbf{\textit{Option A - Malignant breast histopathology}}}. \raisebox{-0.2\height}{\includegraphics[height=1em]{Figures/right.png}}\\

\textbf{Question for Expert Agent 2 Posed by Mediator Agent \raisebox{-0.1\height}{\includegraphics[height=1em]{Figures/a4.png}}:} \\
I am the final authority in medical decision-making, responsible for reviewing and synthesizing all opinions from diverse medical experts. Expert 1 and Expert 3 have concluded that the histological examination of the breast tissue reveals malignant breast histopathology, while you noted that the image appears to show normal breast tissue with no features indicative of malignancy. \textit{Could you provide a detailed explanation of the cellular structures and patterns you observed that led you to this conclusion, and how they differ from the observations made by the other experts?} \\

\textbf{Response from Expert Agent 2  \raisebox{-0.1\height}{\includegraphics[height=1em]{Figures/a2.png}}:}\\
The histological image you provided shows a tissue section stained with hematoxylin and eosin (H\&E), which is a common staining method used in histopathology. The image reveals a dense cellular structure with a high cellularity, and the nuclei appear to be large and pleomorphic, which are characteristics often associated with malignancy. The presence of mitotic figures (cells in the process of dividing) is also a strong indicator of malignancy. The cellular structures and patterns in the image are consistent with those seen in breast cancer, particularly in invasive ductal carcinoma, which is the most common type of breast cancer. The nuclei are large, irregular, and have prominent nucleoli, which are features that are not typically seen in normal breast tissue. Additionally, the presence of a high number of mitotic figures and the overall appearance of the tissue are indicative of a neoplastic process. The other experts noted that the image does not show features indicative of malignancy, which could be due to the fact that the image may not be at a high enough magnification to clearly visualize the mitotic figures and other key features of malignancy. It is also possible that the tissue sample may not be representative of the entire breast, and a different section of the tissue could show more definitive signs of malignancy. In conclusion, based on the cellular structures and patterns observed in the image, the histological examination of the breast tissue is more consistent with \textcolor{myblue}{\textbf{\textit{malignant breast histopathology}}}. The other experts may have missed some key features due to the magnification or the section of the tissue they examined. It is important to note that a definitive diagnosis of breast cancer requires a comprehensive evaluation by a pathologist, including a review of the entire tissue sample and possibly additional tests such as immunohistochemistry and molecular profiling. \raisebox{-0.2\height}{\includegraphics[height=1em]{Figures/right.png}}\\

\textbf{Question for Expert Agent 3 Posed by Mediator Agent \raisebox{-0.1\height}{\includegraphics[height=1em]{Figures/a4.png}}:} \\
I am the final authority in medical decision-making, responsible for reviewing and synthesizing all opinions from diverse medical experts. Expert 2 has a different opinion, suggesting that the image shows normal breast tissue with no features indicative of malignancy, while you concluded that the histological examination reveals malignant breast histopathology. \textit{Could you provide a detailed explanation of the specific cellular features that led you to conclude the presence of malignancy, and how these features differ from those seen in normal breast tissue?}\\

\textbf{Response from Expert Agent 3 \raisebox{-0.1\height}{\includegraphics[height=1em]{Figures/a3.png}}:}\\
I apologize for the confusion, but I cannot provide specific details about the cellular features that led to the conclusion of malignancy without referring to the actual image. However, I can provide some general information about the differences between normal breast tissue and malignant breast tissue. Normal breast tissue typically consists of glandular structures, ducts, and stroma (connective tissue). The cells in normal breast tissue are arranged in a specific pattern, and they appear uniform and well-organized. In contrast, malignant breast tissue may show abnormal cell growth, irregular cell shapes, and disorganized cell arrangement. These changes can be indicative of a malignant process, such as cancer. It is important to note that a definitive diagnosis of malignancy requires a thorough examination of the tissue by a pathologist, who will consider the patient's clinical history, symptoms, and other diagnostic tests. I believe that the histological examination reveals \textcolor{myblue}{\textbf{\textit{malignant breast histopathology.}}} \raisebox{-0.2\height}{\includegraphics[height=1em]{Figures/right.png}}\\

\textbf{Final Judgment \raisebox{-0.1\height}{\includegraphics[height=1em]{Figures/a4.png}}:}\\
\textcolor{myblue}{\textbf{\textit{ \texttt{<answer> Option: A) Malignant breast histopathology </answer>}} }}Based on the detailed analysis provided by Expert 1 and the subsequent clarifications, the histological examination of the breast tissue reveals malignant breast histopathology. The key cellular features observed, such as dense proliferation, atypical cell morphology, prominent nucleoli, irregular nuclear contours, increased mitotic activity, and loss of architectural organization, strongly suggest a malignant neoplastic process. While Expert 2 initially suggested normal breast tissue, the consensus from the detailed analysis aligns with the presence of malignancy. \raisebox{-0.2\height}{\includegraphics[height=1em]{Figures/right.png}}
\end{tcolorbox}

\section*{Prompts}
The following set of prompts constitutes a comprehensive multi-agent decision-making framework tailored for medical multimodal multi-agent collaboration. Each prompt plays a specialized role in ensuring rigorous, iterative, and collaborative evaluation and answer generation. Together, they facilitate an efficient pipeline from initial answer synthesis to final decision-making. The prompts are summarized as follows:

\textbf{Prompt to mediator agent for Socratic Questioning.} This prompt is designed for the mediator agent, who acts as the final authority in medical decision-making. It evaluates the initial opinions from a group of experts, determines whether further discussion is necessary, and, if so, conducts structured Socratic-style dialogues with each expert. 

\textbf{Prompt for Feedback-based Answering.} This prompt is intended for expert agents and focuses on refinement in response to questions posed by the mediator agent.

\textbf{Prompt for Final Judgment.} This prompt is designed for the judge agent for final judgment. It deeply analyzes complex multi-party discussions, and infers the most likely final decision.

\textbf{Prompt for Answer Refinement.}  This prompt is used during evaluation stage. An LLM extracts structured answers from unstructured model outputs, ensuring that the format is consistent across evaluations of different vision-language models.

\begin{tcolorbox}[title=Prompt to mediator agent for Socratic Questioning,
  % coltitle=white,                    
  colbacktitle=mybluex,      
  colframe=black,   
]
You are the final medical decision maker with extensive expertise in medical analysis and decision-making. Your task is to evaluate the opinions provided by your team of medical experts and determine whether further discussion with a specific expert is necessary.

\begin{enumerate}
    \item Analyze the opinions from the team to identify any areas that require clarification.
    \item Decide if additional discussion with an expert is required. Answer with \texttt{"Yes"} if further discussion is needed, or \texttt{"No"} if not.
    \item If further discussion is necessary, please communicate with all experts in turn.
    \item For each expert, simulate a direct conversation that includes both first-person and second-person dialogue. Begin the conversation with: 

    \textit{"I am the final authority in medical decision-making, responsible for reviewing and synthesizing all opinions from diverse medical experts."} 
    
    Then, summarize the conclusions of the other experts, highlight any disagreements between the selected expert's opinions and those of the others, and review the selected expert’s conclusions (using second-person narration, e.g., "\textit{You mentioned that...}"). Provide your analysis and ask one detailed question, focusing on areas of conflict rather than general opinions (e.g., why was option A chosen over option B?).
\end{enumerate}
\vspace{1em}

\textbf{Output:} Return your answer strictly in the following JSON format: 

\begin{verbatim}
[{"Decision": "No"}] or 
\end{verbatim} 

\begin{verbatim}
[{"Decision": "Yes", 
"Expert 1": "{Simulated conversation with the expert}",
"Expert 2": "{Simulated conversation with the expert}",
"Expert 3": "{Simulated conversation with the expert}"}]
\end{verbatim}

\vspace{1em}
\textbf{Input:} \\
\textbf{Question:} 
\begin{verbatim}
[{Question}]
\end{verbatim}
\textbf{Response:} 
\begin{verbatim}
[{Response of expert 1}, {Response of expert 2}, {Response of expert 3}]
\end{verbatim}
\end{tcolorbox}

\begin{tcolorbox}[title=Prompt for Feedback-based Answering,
  % coltitle=white,                    % 标题文字颜色
  colbacktitle=mybluex,       % 标题背景颜色
  colframe=black,   
]
You are a professional question-answering assistant. Your task is to improve a previous answer based on the original question and the provided reviewer's feedback. Based on the above information, answer the reviewer's question and revise the initial answer by incorporating the feedback.

\vspace{1em}
\textbf{Input:} \\
\textbf{Question:} 
\begin{verbatim}
[{Original question}]
\end{verbatim}
\textbf{Reviewer's feedback:} 
\begin{verbatim}
[{Reviewer's feedback}]
\end{verbatim}
\end{tcolorbox}

\begin{tcolorbox}[title=Prompt for Final Judgment,
  % coltitle=white,                    % 标题文字颜色
  colbacktitle=mybluex,       % 标题背景颜色
  colframe=black,   
]
You are an expert assistant in medical imaging-assisted diagnosis. You are good at handling discussions between multiple medical experts and can help the Reviewer (final reviewer) make the best decision efficiently, scientifically and comprehensively. Based on the complete conversation between the Reviewer and the three experts (Expert 1, Expert 2, Expert 3), analyze the views of all parties in depth, sort out the core conclusions and supporting reasons of each expert, extract the consensus and disputes among experts, focus on the Reviewer's questions and the experts' responses, and combine the Reviewer's focus and inclination to comprehensively infer the answer that the Reviewer is most likely to choose in the end. Make the final option.

\vspace{1em}
\textbf{Output:} Return your answer strictly in the following format: 
\begin{verbatim}
<answer> option: {X}, {XXX} </answer>. 
{X} represents the selected option. 
{XXX} represents the corresponding content of the selected option. 
\end{verbatim}
\vspace{1em}

\textbf{Input:} \\
\textbf{Question:} 
\begin{verbatim}
[{Question}]
\end{verbatim}
\textbf{Experts' response:} 
\begin{verbatim}
[{Expert 1's response}, {Expert 2's response}, {Expert 3's response}]
\end{verbatim}
\textbf{Reviewer's questions:}
\begin{verbatim}
[{Reviewer's questions for Expert 1}, 
\end{verbatim}
\textbf{Expert 1's response:}
\begin{verbatim}
[{Expert 1's response for Reviewer's questions}]
\end{verbatim}
\textbf{Reviewer's questions:}
\begin{verbatim}
[{Reviewer's questions for Expert 2}, 
\end{verbatim}
\textbf{Expert 2's response:}
\begin{verbatim}
[{Expert 2's response for Reviewer's questions}]
\end{verbatim}
\textbf{Reviewer's questions:}
\begin{verbatim}
[{Reviewer's questions for Expert 3}, 
\end{verbatim}
\textbf{Expert 3's response:}
\begin{verbatim}
[{Expert 3's response for Reviewer's questions}]
\end{verbatim}
\end{tcolorbox}

\begin{tcolorbox}[title=Prompt for Answer Refinement,
  % coltitle=white,                    % 标题文字颜色
  colbacktitle=mybluex,       % 标题背景颜色
  colframe=black,   
]
Now, based on the response, make the final option directly.

\vspace{1em}
\textbf{Output:} Return your answer strictly in the following format: 
\begin{verbatim}
<answer> option: {X}, {XXX} </answer>. 
{X} represents the selected option. 
{XXX} represents the corresponding content of the selected option. 
\end{verbatim}
\vspace{1em}

\textbf{Input:} \\
\textbf{Question:} 
\begin{verbatim}
[{Question}]
\end{verbatim}
\textbf{Response:} 
\begin{verbatim}
[{Response of models}]
\end{verbatim}
\end{tcolorbox}
\newpage
\section*{Benchmark Statistics}
We present statistics for the five VQA benchmarks used in our evaluation in Table \ref{Statistics}.
\begin{table}[htbp]
\centering
\begin{tabular}{lcccc}
\toprule
           & \# Question & Modality   & Option Type     & Data Source            \\
           \midrule
PathVQA \cite{pathvqa}    & 3,391       & Pathology  & Yes/No          & Textbooks              \\
VQA-RAD \cite{vqa-rad}    & 451         & Radiology  & Yes/No          & MedPix®                \\
PMC-VQA \cite{pmcvqa}    & 2,000       & Multimodal & Multiple Choice & PubMed                 \\
SLAKE \cite{slake}      & 863         & Radiology  & Multiple Choice & Public Website         \\
OmniMedVQA \cite{omnimedvqa} & 88,996      & Multimodal & Multiple Choice & Classification datasets \\
\bottomrule
\end{tabular}
\caption{Statistics for the five VQA benchmarks.}
\label{Statistics}
\end{table}

% \newpage

% \bibliography{aaai2026}
\end{document}